%% file: paper.tex
\documentclass[times,twocolumn,final]{elsarticle}

\usepackage{framed,multirow}

\usepackage{latexsym}

\usepackage{graphicx}
\usepackage{amsmath}
\usepackage{amssymb}
\usepackage{color}
\usepackage[table]{xcolor}
\usepackage[lofdepth,lotdepth, caption=false,font=footnotesize]{subfig}
\usepackage{hhline}
\usepackage{stfloats}
\definecolor{Gray}{gray}{0.9}
\usepackage[utf8]{inputenc}
\usepackage{hyperref}

\input{figs}

\input{tabs}

\newcommand{\eqnref}[1]{Eq.~(\ref{eqn:#1})}
\newcommand{\figref}[1]{Fig.~\ref{fig:#1}}
\newcommand{\tblref}[1]{Table~\ref{tbl:#1}}

\begin{document}


\begin{frontmatter}

\title{Pressure Eye: In-bed Contact Pressure Estimation \\via Contact-less Imaging}%

\author[1]{Shuangjun  Liu}
\author[1]{Sarah Ostadabbas \corref{cor1}}
\cortext[cor1]{Corresponding author:   Email: ostadabbas@ece.neu.edu}

\address[1]{Augmented Cognition Lab, Department of Electrical and Computer Engineering, Northeastern University, Boston,
MA, USA.}


\begin{abstract}
Computer vision has achieved great success in interpreting semantic meanings from images, yet estimating underlying (non-visual) physical properties of an object is often limited to their bulk values rather than reconstructing a dense map. In this work, we present our pressure eye (PEye) approach to estimate contact pressure between a human body and the  surface she is lying on with high resolution from  vision signals directly. PEye approach could ultimately enable the prediction and early detection of pressure ulcers in bed-bound patients, that currently depends on the use of expensive pressure mats. Our PEye network is configured in a dual encoding shared decoding form to fuse visual cues and some relevant physical parameters in order to reconstruct high resolution pressure maps (PMs). We also present a pixel-wise resampling approach based on Naive Bayes assumption to further enhance the PM regression performance. A percentage of correct sensing (PCS) tailored for sensing estimation accuracy evaluation is also proposed which provides another perspective for performance evaluation under varying error tolerances. We tested our approach via a series of extensive experiments using multimodal sensing technologies to collect data from 102 subjects while  lying on a bed. The individual's high resolution contact pressure data could be estimated from their RGB or long wavelength infrared (LWIR) images with 91.8\%  and 91.2\% estimation accuracies in $PCS_{efs0.1}$ criteria, superior to state-of-the-art methods in the related image regression/translation tasks. 

\end{abstract}

\begin{keyword}
 Bedsores \sep Contact pressure estimation \sep Multimodal in-bed pose data \sep Image generation \sep Pressure ulcers \sep Style translation.
\end{keyword}

\end{frontmatter}

\section{Introduction}
\label{sec:intro}
In the past few years, advances in the field of computer vision have been actively introduced into the healthcare and medical fields including in-bed (or at-rest) behavior studies, such as works related to  sleep pose  estimation \cite{clever20183d,liu2019bed,liu2019seeing,liu2017vision}, sleep quality assessment \cite{samy2013unobtrusive,mendoncca2019review,febriana2019sleep} or in-bed posture and action recognition \cite{ostadabbas2014bed,martinez2015action,liu2014breathsens}. One area that significantly benefits from automatic in-bed behavior monitoring is pressure ulcer (also known as bedsore) prevention and early detection studies in bed-bound patients. Pressure ulcers are localized injuries to the skin and/or underlying tissues usually over a bony prominence as a result of prolong contact pressure caused by the bed surface \cite{black2007national}. Pressure ulcers are extremely costly to treat and may lead to several other health problems, including death. Currently, the in-bed contact pressure monitoring systems mainly rely on the use of commercially available pressure mats, which are expensive and hard to maintain \cite{ostadabbas2011pressure}. In this work, we present our contact-less pressure eye (PEye) approach that employs vision signals to infer high resolution  contact pressure between human body and its lying surface.

\comparison


%
Perceiving non-visual physical properties of an object in an image is a subject recently explored by many work, such as finding total manipulation force due to hand-object interactions \cite{pham2015towards},  inferring force distribution along elastic objects \cite{greminger2004vision}, and estimating body weights \cite{velardo2010weight}. Yet, the majority of these works  focus on estimating the bulk physical properties such as weight or total force, rather than providing a dense map of the physical quantity of interest. Visual signals are often employed for displacement measurement purposes in these works, while problem solving still heavily relies  on physical modeling  including kinematics, dynamics, or elasticity theories. Using PEye approach, we aim at recovering a dense map of in-bed contact pressure based on its correspondence to the vision field. 

We take into consideration  the main factors in the pressure forming process, not only  the visual cues  but also  pressure-related physical quantities as our inputs and regress them directly to estimate the contact pressure. By employing the deep neural network learning capabilities, PEye  avoids the sophisticate mechanical modeling of the contact pressure and solves the problem in an end-to-end manner. We form the PEye network with multi-stage dual encoding, shared decoding structure with both visual and non-visual physical inputs, and enforce physical law during supervision. Besides the conventional visual modality, the RGB, we also evaluate our PEye approach when the vision signal is based on the long wavelength IR (LWIR) imaging. LWIR 
stays effective consistently even under complete darkness, which enables long-term in-bed behavior monitoring throughout the day with varying/none illumination states.  
 



\subsection{Related Work}
In healthcare domain, there have been several recent attempts to infer underlying non-visual physical/physiological quantities from accessible and unobtrusive vision signals. In \cite{poh2010advancements}, heart rate, respiration rate, and heart rate variability are measured via a webcam. Authors in \cite{ostadabbas2015vision} used a Microsoft Kinect
depth camera to measure  lung function and
the severity of airway obstruction in  patients with active obstructive pulmonary diseases. In \cite{murthy2009thermal}, during polysomnography, thermal imaging was employed to monitor the airflow to detect potential apnea and hypopnea. 
The thermal modality has also been introduced for high temperature (i.e. fever) detection from faces in \cite{nguyen2010comparison}.


Among the medically-relevant signals that can be captured using unobtrusive sensing technologies, contact pressure sensing also plays an important role for bed-bound patient  monitoring including the prevention and early detection of pressure ulcers  \cite{ostadabbas2012resource}. Pressure ulcers are localized injuries to the skin and/or underlying tissues usually over a bony prominence as a result of prolong contact pressure caused by the bed surface \cite{black2007national}. Pressure ulcers are extremely costly to treat and may lead to several other health complications, including amputation or even death. Currently, the in-bed  pressure monitoring systems mainly rely on the use of commercially available pressure sensing mats, which are expensive and hard to maintain \cite{ostadabbas2011pressure}. 
Yet, there is no existing work to estimate dense pressure map (PM) directly from a vision signal.

Nonetheless, the problem of contact pressure estimation  in abstract shows high similarity to some recent computer vision works. Given an image $X$ in domain $A$, inferring its underlying physical quantity map, $Y$ in domain $B$ can mathematically be represented as a dense regression problem, $Y=\phi(X)$. 
There are existing computer vision tasks that are formulated in this fashion or a similar way either by focusing on regression accuracy or visual appeal, as shown in \figref{comparison}. A well-studied example is  human pose estimation, where $X$ is an RGB image from the human body and $Y$ is the body joint confidence map (see \figref{comparison}(a)). Particularly in this problem, the joint coordinate regression problem is transferred to a maximum localization problem of the confidence maps which is regressed in a dense manner \cite{cao2018openpose,wei2016convolutional}.  A similar yet different application is semantic segmentation \cite{long2015fully}, in which $X$ is an RGB image and  $Y$ is its semantically segmented image, which could be seen as a pixel-wise classification problem.
Another familiar example of dense regression is image restoration (see \figref{comparison}(b)), where high quality images are recovered from degenerated ones, including de-noising, super resolution, and de-blocking \cite{tai2017memnet}. Besides exact regression, estimating (generating) a new image from a given image/map is also similar to image translation  task via generative adversarial networks (GAN) oriented models \cite{zhu2017unpaired}, where instead of exact recovery of the ground truth, they seek for visual pleasantness of the translated image with preferred ``styles'' (see \figref{comparison}(c)). 

Given the existing body of work, one may conclude that
PM estimation from vision is a solved problem based on its regression-based nature. 
However, the following concerns still exist. First, having an image/map $X$ is not always sufficient to estimate a physical quantity $Y$.  In our working example, suppose we have two similar size/color cubes on a surface but with different weights (e.g. iron vs. plastic). While, they look visually similar, their contact pressure is vastly different. Therefore, additional physical property is needed to solve the problem, for example providing the material types.  In that case, the inference is based on both visual and non-visual signals, instead of $X$, alone. 
Secondly,  while at  first glance PM looks just like  a style-differed image (see \figref{comparison}(d)), there is a ``long'' domain difference between the visual input  and the non-visual PM. Looking closer at the \figref{comparison}(d), it is apparent  that although $X$ and $Y$ have similar pose, they are not sharing the same profile. 

Moreover, existing works are proposed under their specific contexts, and they have never been tested in a contact pressure estimation context.  In the case of human pose estimation, confidence map regression is just a byproduct of the algorithm and not the end goal and the correct maximum localization is not equivalent to the quality of the generated heatmap itself, which is a surjective mapping. In image restoration examples, the inference is based on degenerated data in the same domain that is  very similar to original high quality image, which means the distance between $X$ and  $Y$ domains is short. This characteristic is  well-employed for problem solving in related studies, where instead of a total remapping, image restoration can be solved more effectively in a residue learning process \cite{tai2017memnet,zhang2018residual}. Even in the case of models such as GeoNet, which is a dense regression of depth and optical flow from RGB sequence \cite{yin2018geonet}, a depth image is still a neighboring domain of its RGB image counterpart and shares strong profile with the input frame, which is employed in the GeoNet model  supervision. However, to estimate dense non-vision physical properties from an image, the mapping is often between two fully distinct domains that differ not only in patterns but over all profiles. Moreover, GAN-based image generation approaches do not pursue pixel-wise accuracy and the discrimination loss is  only based on visual plausibility of the generated maps and even encourage diversity in this process \cite{salimans2016improved}. Such $1$-to-$n$ mapping process essentially contradicts with a unique regression ($1$-to-$1$ mapping) purpose.



\PE
\subsection{Our Contributions}
In contrast to the above mentioned works,  pressure map (PM) estimation is a regression problem, in which pixel-wise accuracy is required. Moreover, due to its medical``decision support'' nature,  the distribution patterns should be presented accurately in order to correctly identify high pressure concentration areas. Compared to a direct translation, PM estimation is more like a data-driven solver that takes the visual clues from RGB or LWIR images and solves the contact pressure mechanics problem in an end-to-end manner with high resolution. Given the specific context and concerns around the PM estimation task, this paper presents PEye approach and explores the possible improvements over the existing regression-based approaches to achieve  better in-bed contact pressure estimation. We summarize the contributions of this paper as follows\footnote{The PEye code  can be found at GitHub: \href{https://github.com/ostadabbas/PressureEye}{Pressure Eye}.}:

\begin{itemize}
    \item Design and implementation of the PEye network in a dual encoding shared decoding form to fuse the visual cues and the relevant physical measures in order to reconstruct a high resolution PM of a person in a given in-bed position. 
    
    \item Based on a Naive Bayes assumption, a pixel-wise resampling (PWRS) approach is proposed to enhance the peak value recovery in sparse PM profiles. 

    \item An evaluation metric to measure pressure sensing performance. As standard metrics for  regression problems, either mean square error (MSE) or peak signal-to-noise ratio (PSNR) only provide an overview of the reconstruction results, here we propose the percentage correct sensing (PCS) metric, which provides another perspective on how a ``sensing array'' performs under varying error tolerance.

    \item Evaluating PEye performance on the largest-ever in-bed multimodal human pose dataset with $>$100 human subjects with nearly 15,000 pose images called  Simultaneously-collected multimodal Lying Pose (SLP), including RGB, LWIR, and contact pressure map, which is publicly available in our webpage \cite{liu2020simultaneously}. 
\end{itemize}

 \section{Pressure Eye (PEye) Approach}
\label{sec:PADS}
Our PEye approach takes RGB or LWIR images as its source domain and generates pressure map (PM) signals as its target domain, as shown in \figref{PE}.
Estimating the contact pressure  using only a vision signal is an ill-posed problem, as RGB or LWIR cannot correctly represent pressure maps of objects with similar appearance but different densities. Furthermore, the differences in human body shapes/geometries lead to different pressure distributions on bed. Therefore, besides a vision signal (RGB or LWIR), we  feed a  physical vector $\beta$ into the PEye network, which incorporates several physique parameters from our human participants including their body weight, gender, and their tailor measurements. 

\PEmodule

PEye network has a multi-stage design by stacking multiple building modules. Each module has a skipped encoding-decoding structure, which is similar to the building block employed in \cite{wei2016convolutional,tai2017memnet,newell2016stacked}. To feed in the physical vector $\beta$,  additional reconfiguration is applied, such that a physical encoder is added to form a dual encoding, shared decoding structure.
The vision signal $X$ and the physical measures $\beta$ are encoded separately, then concatenated and decoded jointly as shown in \figref{PEmodule}. The decoding process turns out to be based on the merged code of both vision and physical vectors.  The convolution and deconvolution layers are both followed by the reLU and batch normalization layers. 

Looking at the PM samples in our dataset, compared to an RGB image, PM looks quite different in style.  Specifically, PM looks more simple, unfilled, with several hollow areas. PM  is also unevenly distributed and concentrated only on a very few supportive areas such as hips and shoulders that generate  sharp peaks. Therefore, many PM pixels are in low value range with only very few in high value range. This can be quantitatively reflected in the histogram comparison of an RGB image and its PM counterpart, as shown in \figref{hist}.  An immediate question is if the PM value distribution imbalance would affect the  performance of the pressure regression task.

\subsection{Pixel-wise Resampling (PWRS)}
\label{sec:PWRS}
Data imbalance issue is a long lasting topic in machine learning (ML) community for classification problems \cite{japkowicz2000class}.  In a binary classification problem, class imbalance occurs when one class, the minority group, contains significantly fewer samples than the other class, the majority group \cite{johnson2019survey}. In many problems \cite{rao2006data,wei2013effective,herland2018big}, the minority group is the class of higher interest, i.e. the positive class. This is also true in our case, in which  we care more about the high pressure concentrated areas which turns out to be the minority as  \figref{hist}(b) histogram shows. 
Although, most existing methods for imbalanced data  focus on the classification problems \cite{chawla2002smote,kubat1997addressing,batista2004study}, some of their ideas can be  introduced into a regression task, such as the resampling strategy. The purpose of resampling is to make each ``class'' presented in the task with equivalent quantities during training. One intuitive way is oversampling the underrepresented class by multiple times to make each class relatively equivalent in size.  

\figHist

To do this in our case, one immediate solution is  to collect PMs with more readings within high range values. However, in the context of in-bed pressure mapping, this is challenging since high pressure values are mainly concentrated in a small number of supporting areas. 
To simplify this problem, similar to Naive Bayes approach \cite{bishop2006pattern}, we assume the PM pixel values are independent. So, the PM resampling is simplified into the pixel-wise resampling  (PWRS) with a trade-off between the accuracy and sampling feasibility. As the resample number (RSN)  depends on the specific PM pixel value $y(i,j)$ at $i$ and $j$ coordinates, we define a function  $s(y(i,j))$, which maps the pixel value into the corresponding RSN.  
A typical $L_2$ loss after resampling becomes:

\begin{align}
       L_{2}^{pwrs} & = \sum_{i=0}^M \sum_{j=0}^N \sum_{k=0}^{s(y(i,j))}(\hat{y}(i,j) - y(i,j))^2 \\ \nonumber
     & = \sum_{i=0}^M \sum_{j=0}^N s(y(i,j))(\hat{y}(i,j) - y(i,j))^2,
\end{align}
where $M$ and $N$ stands for the row and column size of the PM, and $\hat y$ stands for the estimated PM result. If we simply deem the $s(y(i,j)$ as weight, it is not necessary to be an integer and the resampling of each pixel will be a pixel-wise weighted $L_2$ loss. 
One intuitive way to build RSN function $s$ is making it inversely proportional to the density function of the pixel value $y$, such that  $s(y) = \lambda_{L_2}/p(y)$, where  $\lambda_{L_2}$ is a constant and $p(y)$ is the density function of $y$. Then, the PWRS loss is formulated as:

\begin{equation}
    L_{2}^{pwrs} = \lambda_{L_2} \sum_{i=0}^M \sum_{j=0}^N (\hat{y}(i,j) - y(i,j))^2/p(y(i,j)).
\end{equation}

The PM pixel values in high pressure range are highly sparse. One typical solution  is using Laplace smoothing by adding evenly distributed additional ``hallucinated'' examples. We add a hallucinated weight $\xi$ instead to enhance under weighted pixel values and have a Laplace smoothed loss as:
\begin{equation}
    L_{2-l}^{pwrs} = \lambda_{L_2} \sum_{i=0}^M \sum_{j=0}^N (\hat{y}(i,j) - y(i,j))^2(1/p(y(i,j) + \xi)).
\end{equation}

\subsection{PEye Network Optimization}
For PEye network training, in addition to the $L_{2-l}^{pwrs}$ loss, we introduce the physical loss $L_{2}^{phy}$, which incorporates the dominant law in pressure forming process. Contact pressure is a sophisticated interactive process between the body parts and the support surface. Detailed modeling of this process is complex and against the end-to-end purpose of a data-driven approach. So, we employ the simple but dominant law for pressure forming as:
\begin{equation}
\label{eqn:phy}
L_{2}^{phy} = (c \sum_i \sum_j \hat{y}(i,j) - w_b)^2,
\end{equation}
where $w_b$ stands for the person's body weight, and  $c$ is the contact area  with the bed represented by each PM pixel. This loss reflects a physics principle that integration of pressure over the contact area should be equal to the person's total weight. As a part of the physical vector $\beta$, $w_b$ is included in both input  and the loss function, which inherently shows the network how to utilize the additional physical information. With a dual encoder input, the decoder net is  supervised from both visual and physical perspectives and the total loss function is given as:

\begin{equation}
    \label{eqn:objective}
    L^{total} =  \lambda_{pwrs} L_{2-l}^{pwrs} + \lambda_{phy} L_{2}^{phy},
\end{equation}
where $\lambda_{pwrs}$ and $\lambda_{phy}$  stand for the weights applied to each loss term, respectively. Additional losses can be introduced to further enhance the visual plausibility of the generated maps, such as patchGAN \cite{ledig2017photo} by adversarial learning, or structural similarity index (SSIM) loss \cite{wang2004image}. We will further evaluate their effects on PM reconstruction performance in our ablation study. 

\subsection{Percentage of Correct Sensing  (PCS) Metric}
\label{sec:PCS}
A classical way to evaluate a regression problem is calculating the overall mean square error (MSE), in which
each pixel contribute evenly in the MSE calculation. However, in many sensing applications, the estimation accuracy in active sensing areas is much more important than the irrelevant background.
In this case, MSE of effective area makes more sense, where we only focus on the active sensing area. Inspired by this as well as the probability of correct keypoint (PCK) metric  in human pose estimation models \cite{andriluka20142d}, we further propose a percentage of correct sensing  (PCS) metric of effective area to provide another evaluation under varying error tolerance. 
We define effective PCS metric as: 
\begin{equation}
    \label{eqn:PCS-efs}
    PCS_{efs(\epsilon)} = \frac{|E[efs]<\epsilon|}{|E[efs]|},
\end{equation}
where $E$ is the error map, which is the difference between the ground truth and the estimated map, $|.|$  is an element-wise counting operation (i.e. cardinality), and $efs$ indicates a selection matrix, which  specifies the effective sensing area.  Threshold $\epsilon$ could be set by a domain-specific absolute value or a normalized value based on the sensing range. 

The idea behind PCS comes from the fact that for a physical sensor, as long as its estimated value is within a predefined application-specific tolerance range, it could be assumed as  a qualified  reading. In an array format, we only need to calculate how many sensors are within this range to evaluate their performance. Otherwise,  a few strong outliers can contribute substantially to a high MSE, while most of the estimated values are correct.
PCS also provides a comprehensive view of the sensing performance with varying error tolerance, since  different application scenarios could hold different threshold for errors. In PEye, we chose the $efs$ threshold to be 5\% or 10\% of the maximum value of the PM as the low response pixels  often are the unoccupied areas or the ones of low interest in practice. 

\section{Experimental Analysis}

\subsection{Multimodal Dataset Collection}

To evaluate PEye approach effectiveness in generating  pressure data from  vision signals, in our previous work \cite{liu2020simultaneously}, we formed and publicly released the Simultaneously-collected multimodal Lying Pose (SLP) dataset, where RGB, LWIR and PM signals are collected using a Logitech webcam, a FLIR IR camera,  and a Tekscan pressure sensing mat, respectively.  We collected data from 102 subjects that were instructed to lie down on a twin-size bed and take random poses in natural ways. To encourage pose diversity, participants were instructed to evenly give 15 poses under three rough posture categories as supine, left side, and right side. Data from 12 subjects were left out for test purpose, while the rest were used for PEye network training.  During data collection, we used a weighted marker to achieve a rough cross domain data alignment  via a homography mapping \cite{hartley2003multiple}.

\subsection{Evaluation Metrics}
As a regression problem, our goal is to generate accurate dense maps of contact pressure. Therefore, we report both MSE over effective area $MSE_{efs}$ as well as our proposed $PCS_{efs}$ metrics. To provide a comprehensive evaluation, we also employed popular metrics from related tasks (some of these tasks are shown in \figref{comparison}). For example, in image restoration task, PSNR and SSIM scores are commonly used \cite{tai2017memnet,wang2004image}.

\subsection{Details on Network Training} 
The loss weights $\lambda_{pwrs}$ and $\lambda_{phy}$  are set to 100 and 1e-06 when employed, otherwise to 0.  In our ablation, we also studied the effect of the featured components in relevant tasks such as the discriminator for adversarial learning and SSIM score loss, in which  they are given weight of 1 and 10 respectively, when used. For configuration with discriminator, we employed a 3-layer PatchGAN  structure as presented in \cite{ledig2017photo}. 

Our input and output data are normalized to $[0,1]$ interval according to their dynamic range. 
For each network configuration, we used 30 epochs and 0.0002 learning rate with Adam solver \cite{kingma2014adam}. For the last 5 epochs, learning rate was linearly decayed. As suggested in \cite{brock2018large}, we employed largest batch size based on the available computational resources during training, which was 70 for the PEye network. All models were implemented under pyTorch framework. Training and testing were conducted on a single NVIDIA V100 GPU. 

\tblAblaRGB
\tblAblaIR
\subsection{Ablation Study}
Here, we explore how PWRS and the introduced physical constraint affect the PEye model's performance. It is also interesting to investigate how the featured components/supervision in other similar tasks affect PEye performance.  
This includes the adversarial training strategy (represented as $L_{D}$ loss) for realistic image generation and the SSIM score for structural similarity (represented as $L_{ssim}$ loss).  

We first implemented a model with all necessary inputs including the visual image  $X$ and the physical vector $\beta$ containing body weight (based on \eqnref{phy}). 
We also employed the conventional $L_2$ reconstruction loss during the supervision, which specifically focuses on minimizing the MSE. This first model called ``base'' forms a typical regression problem similar to the most regression tasks focusing only on a $L_2$ loss, for example the human pose estimation \cite{sun2019deep,newell2016stacked}. 
Based on this, we gradually added proposed components including PWRS strategy, body weight constraint, and also $L_{D}$ and $L_{ssim}$ losses to study how they affect the model's performance. In our ablation study, we evaluated proposed components individually as well as jointly. The same ablation is conducted for both RGB and LWIR respectively as the input $X$.

Performance of tested configurations are given in \tblref{ablaRGB} and \tblref{ablaIR} for RGB and LWIR input domains, respectively, where base indicate the base model, pwrs, phy, ssim, D indicates the PEye network with inclusion of PWRS strategy, physical constraint, SSIM loss and discriminator loss with adversarial learning, respectively. The combination of approaches is concatenated with ``-''.
The corresponding PCS plot is shown in \figref{abla}. 
In both tables, it is apparent that although a typical $L_2$ regression strategy in the base model gives decent PSNR, but it shows worst performance for active area recovery as indicated by both $MSE_{efs}$ and $PCS_{efs}$ metrics. Proposed PWRS supervision greatly enhances the active area recovery. Introducing physics law constraint also enhances the performance both individually or combined with the PWRS approach. 
There are also some other interesting findings in our ablation study: (i) MSE-only focused model (i.e. base) does not end up with best MSE/PSNR performance, and (ii) although not specially focused on active area, local structure regulation (SSIM and discriminator losses) or physical constraints both enhance model performance in the active area. We believe $L_2$ regression by itself can easily get stuck at local optimal by focusing only on average performance. Additional constraints however can push it out of the comfort zone towards optimum. 
Please note that adding discriminator (D) loss or SSIM loss to the pwrs-phy though weakens regression oriented metrics such as $PCS_{efs}$  and $MSE_{efs}$, yet is in favor of the SSIM score and visual satisfaction. This may be due to the fact that $L_D$, $L_{ssim}$, and $L_{2-l}^{pwrs}$  all play a role for the active area pattern forming, which counter each other's effect. Yet, $L_{2-l}^{pwrs}$ is a sensing focused loss and is favored more by the regression oriented metrics. 

\tblstg
\figAbla

\textbf{Multi-stage Setting:} We also conducted analysis to evaluate the  effect of the multistage structure on model performance of pwrs-phy configuration, as shown in \tblref{stg} . The results demonstrate that additional stacked modules improve $PCS_{efs}$ metric slightly compared to the single stage configuration. However adding additional stages after 2 does not show obvious improvements, as the major refinements are bound to happen in the early stages. A recovery example with different stages of PEye is also shown in \figref{stgsRGB}, from which we can see with single stage structure, the recovery has more exaggerated and more pressure areas such as the torso area. With additional stages, the pattern becomes more slim and clean with higher SSIM and PSNR score. 

\tblBetaB
\stgsRGB

\textbf{Physical Parameters:} The detail of the physical parameters of $\beta$ is shown in \tblref{betaB}. 

In the PEye approach, body weight is assumed to be the dominant factor for contact pressure generation. To further investigate the effect of other physical parameters for PM data reconstruction, we also collected participants' height, gender, and the tailor measurements of all of their major limbs (head, legs, and arms) and torso parts (bust, waist, and hip), which are listed in \tblref{betaB}. For gender, 0 is used for female and 1 for male. All limb measurements are from right side with the body symmetry assumption. These parameters are added gradually to the $\beta$ in addition to the weight parameter, where the tailor measurements from 4th to 10th entries are added together. $PCS_{efs}$ performance with varying length of $\beta$ are shown in \figref{phyNum} for both LWIR and RGB input modalities. 

In order to  illustrate the differences better, zoomed-in version of both \figref{nPhy_RGB} and \figref{nPhy_IR} are shown in \figref{nPhy_RGB_zoomIn} and \figref{nPhy_IR_zoomIn}, respectively.
Basically, additional physical inputs will improve the performance, but not always for a specific measure. For instance, while adding gender shows positive effect when LWIR is the input domain, yet it is not the case when RGB images are used as the input. Full $\beta$ version however gives the best performance in both domains.  
We believe this is reasonable, as extended $\beta$ is supposed to provide more details of person's body shape and physiques, which are relevant factors in pressure map forming. 

\figPhyNum

\figGridsSht

Qualitative comparison of the generated pressure maps from different PEye network configurations are also shown in \figref{gridsSht}. 
Due to the space constraints, we only listed typical ablation configurations for qualitative comparison. We employed same sample pair from both RGB and LWIR domains for an easier comparison across domains.
In supine pose, base model is susceptible to ignoring pressure areas with small values such as arms, which are well-preserved in the pwrs-phy version. With additional SSIM and D supervision, the generated patterns are more visually pleasant. In a closer look, even the high frequency jitters effects of the raw measurements are well-preserved. However, overall estimated pressure is prone to be lower than ground truth. High pressure points are not always preserved, such as the peak areas around the elbows in the supine position and heel pressure area is shrunk. These results once more reflect the different focus between our PM regression approach and the conventional image generation tasks. Image generation tasks usually focus on the generated patterns rather than the  absolute regression accuracy, while in PEye, both patterns and the regression performance are important for an accurate sensing inference. Referring to the qualitative result in \figref{gridsSht}, we also found out that visually superior images usually hold better effective PCS performance, which supports the feasibility of the PCS metric for this task.  

 
Overall, PM reconstruction performance is similar in both RGB and LWIR domains and their metric performance are similar. We believe that although LWIR loses much of the appearance details yet the human body profile is still clearly preserved for semantic recognition of different body parts, which is required for contact pressure estimation.

\figSota

\subsection{Comparison with the State-of-the-Art}
As the PEye dense PM regression task is proposed here for the first time, there are no other studies for an exact comparison. Instead, we chose representative methods (shown in \figref{comparison}) from similar tasks either for regression goal or the image translation goal. From  the problem formulation perspective, as a dense regression problem, human pose estimation (e.g. openPose \cite{cao2018openpose}) and image restoration (e.g. memNet \cite{tai2017memnet}) tasks  are both very similar to ours. From an image translation perspective, our problem can also be described as generating maps in one domain using data in another domain. This task is also very similar to the image translation task and can be conducted with or without pairwise correspondence. So we chose pix2pix \cite{isola2017image} and cycleGAN \cite{zhu2017unpaired} as representative models, respectively. 

\tblSotaRGB
\tblSotaIR

We adapted these models for our task with ``as it is'' principle to minimize side effect of unnecessary changes. OpenPose holds two branches of part affinity field (PAF) and joint confidence map (heat map). Since PAF is not available in our case, we only kept the confidence map branch with pressure map as its ground truth for the supervision. Following its official implementation, the full resolution heat map is recovered via bilinear interpolation.  As for memNet, it  relies on residue learning mechanism which requires identical data formats in input and output. We fed the network with gray scale image to match the single channel pressure map. Both pix2pix and cycleGAN adaptation were straightforward by replacing the input and output with our data. 

The comparison with the state-of-the-art is conducted in both RGB and LWIR domains as reported in \tblref{sotaRGB} and \tblref{sotaIR}, respectively with PCS plot shown in \figref{sota}. 
In $MSE_{efs}$ and $PCS_{efs}$ metric, pwrs-phy still shows noticeable improvements over  other methods. OpenPose also shows a good performance especially in metrics such as PSNR and SSIM score. Referring to the qualitative results in \figref{gridsSht}, we can see that openPose presents a good overall pattern by localizing the high pressure areas most times correctly. However,  these peaks are usually not high enough compared to the ground truth 
and the whole map has ended up to be soft and blurry. Low pressure or small areas such as arm areas are  also prone to be ignored. These results agree with our discussion earlier that human pose estimation focuses on confidence map as a byproduct for pose regression. Though effective to localize  possible pressure concentration areas yet it is not focusing on the regression accuracy but relative correctness.  

MemNet basically fails to learn anything in RGB domain and also performs poorly in LWIR domain. This may be due to the fact that image restoration is built on top a quite similar input data from the same domain, where residue learning is reasonable. However, our task does not provide such near neighbor convenience. 
pix2pix   provides nice details in local areas, yet overall it is prone to yield lower response than ground truth. Small parts such as foot and hand area are also sometimes missing in the recovery . 
CycleGAN only yields partially recovery with distorted results. We believe point-wise supervision between correspondence is quite important for the regression purpose, which cycleGAN lacks. CycleGAN shows better recovery with LWIR input than RGB counterpart. This may be caused by the domain similarity between PM and LWIR, in which  the body part areas highlighted by high temperature in LWIR correspond to high pressure areas in PM. We can also see that though high SSIM score is achieved by cycleGAN, the reconstruction is not necessarily satisfactory.

\section{Conclusion}
In this study, we explored the possibility of recovering the contact pressure between a lying human and the bed surface from a vision signal (RGB or LWIR) in a dense regression manner. Our PEye approach can potentially lead to a  cost-efficient high resolution pressure mapping, since we were able to  recover expensive pressure map signals from  low cost vision signals. A large-scale in-bed pose dataset is also formed and  released that contains  simultaneously collected multi-domain data from human while lying in bed, with large  enough size to train deep neural networks from scratch. We conducted PEye approach evaluation with RGB and LWIR as source domains and compared its performance extensively across similar tasks with their state-of-the-art models. From the comparison, we also found out that although formulated exactly the same way, every task holds its specific context and focus, in which a well-performed model for one task does not necessarily guarantee optimal performance in another task. 
In our evaluation, we  noticed that when using PEye approach some failure cases with fake pressure in  unsupported areas may appear. This usually happens when body parts are partially supported by each other and not the bed and the network fails to recognize such elevation and reports false pressures. Addressing this challenge is the topic of  our future study. 

\section{Acknowledgement} 
This study was supported by the National Science Foundation (NSF-IIS 1755695). We would also like to thank Xiaofei Huang, Zhilan Li, Zhun Deng, Cheng Li and Guannan Dong for their contribution in SLP data collection.  
\bibliographystyle{model2-names.bst}\biboptions{authoryear}
\bibliography{paper}



\end{document}

%% file: figs.tex
\newcommand{\comparison}{
\begin{figure*}[t]
    \centering
    \includegraphics[width=.95\linewidth]{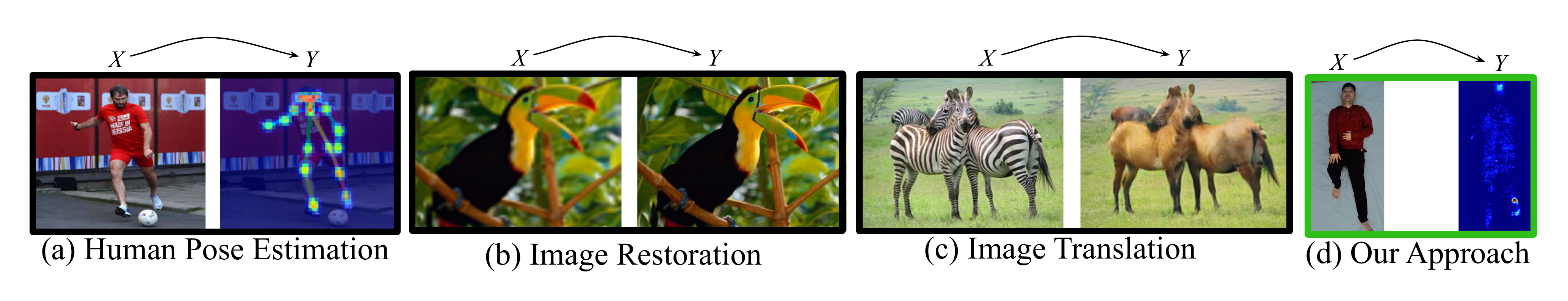}
    \caption{Comparing the existing image generation tasks: (a) human pose estimation  \cite{cao2018openpose}, (b) image restoration  \cite{tai2017memnet}, and (c) image translation \cite{zhu2017unpaired}, with our (d) pressure eye (PEye) approach for in-bed contact pressure estimation.}
    \label{fig:comparison}
\end{figure*}
}

\newcommand{\PE}{
\begin{figure*}[t]
    \centering
    \includegraphics[width=.95\linewidth]{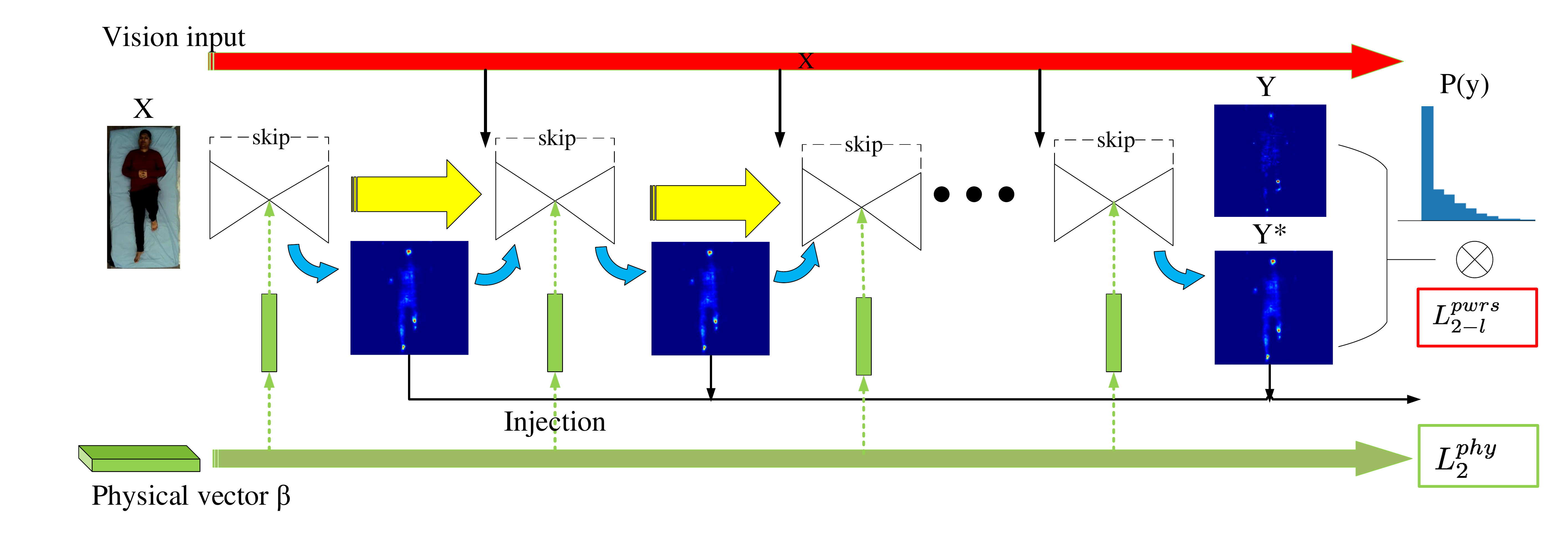}
    \caption{Pressure eye (PEye) network structure for in-bed  contact pressure estimation.}
    \label{fig:PE}
\end{figure*}
}

\newcommand{\PEmodule}{
\begin{figure}[t]
    \centering
    \includegraphics[width=0.95\linewidth]{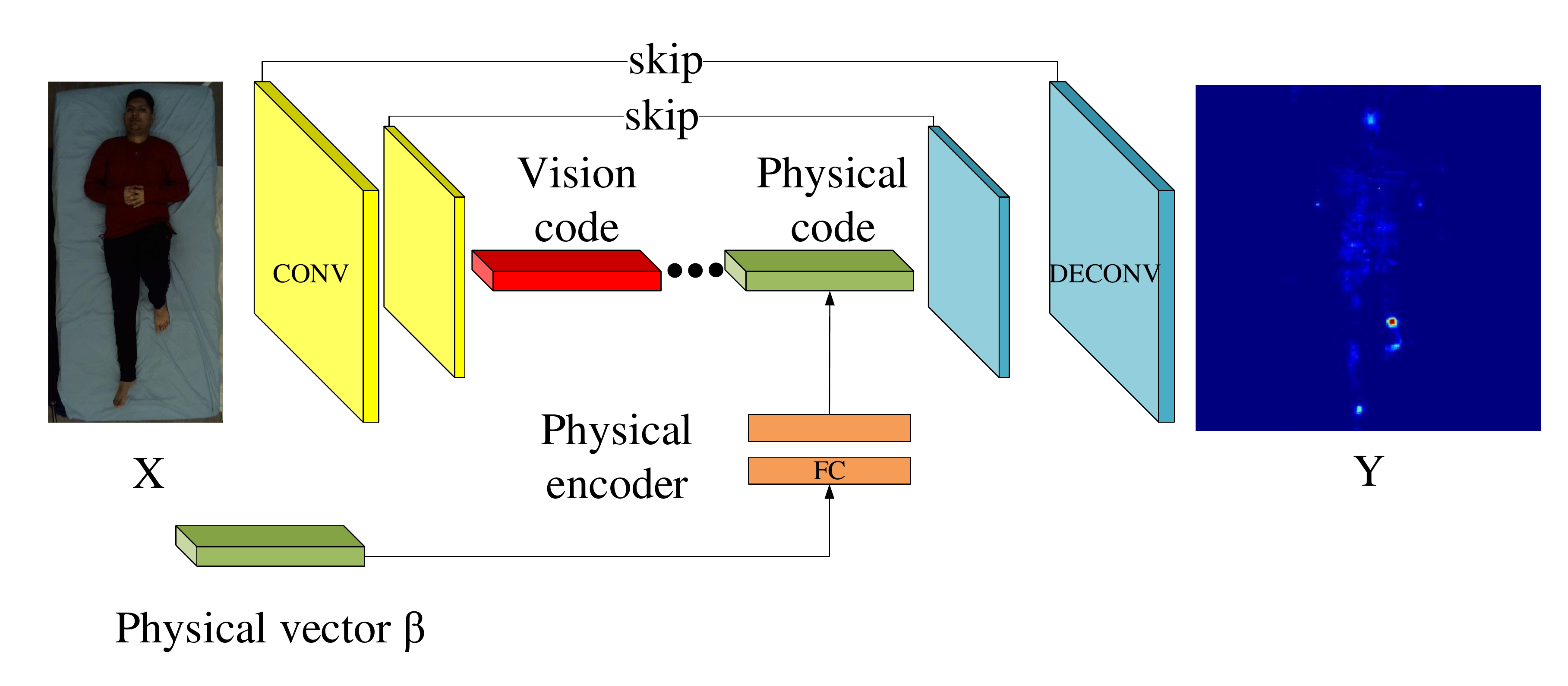}
    \caption{An overview of the PEye network building blocks.}
    \label{fig:PEmodule}
\end{figure}
}

\newcommand{\figHist}{
\begin{figure}[t]
    \centering
\subfloat[]{
    \includegraphics[width=0.45\linewidth]{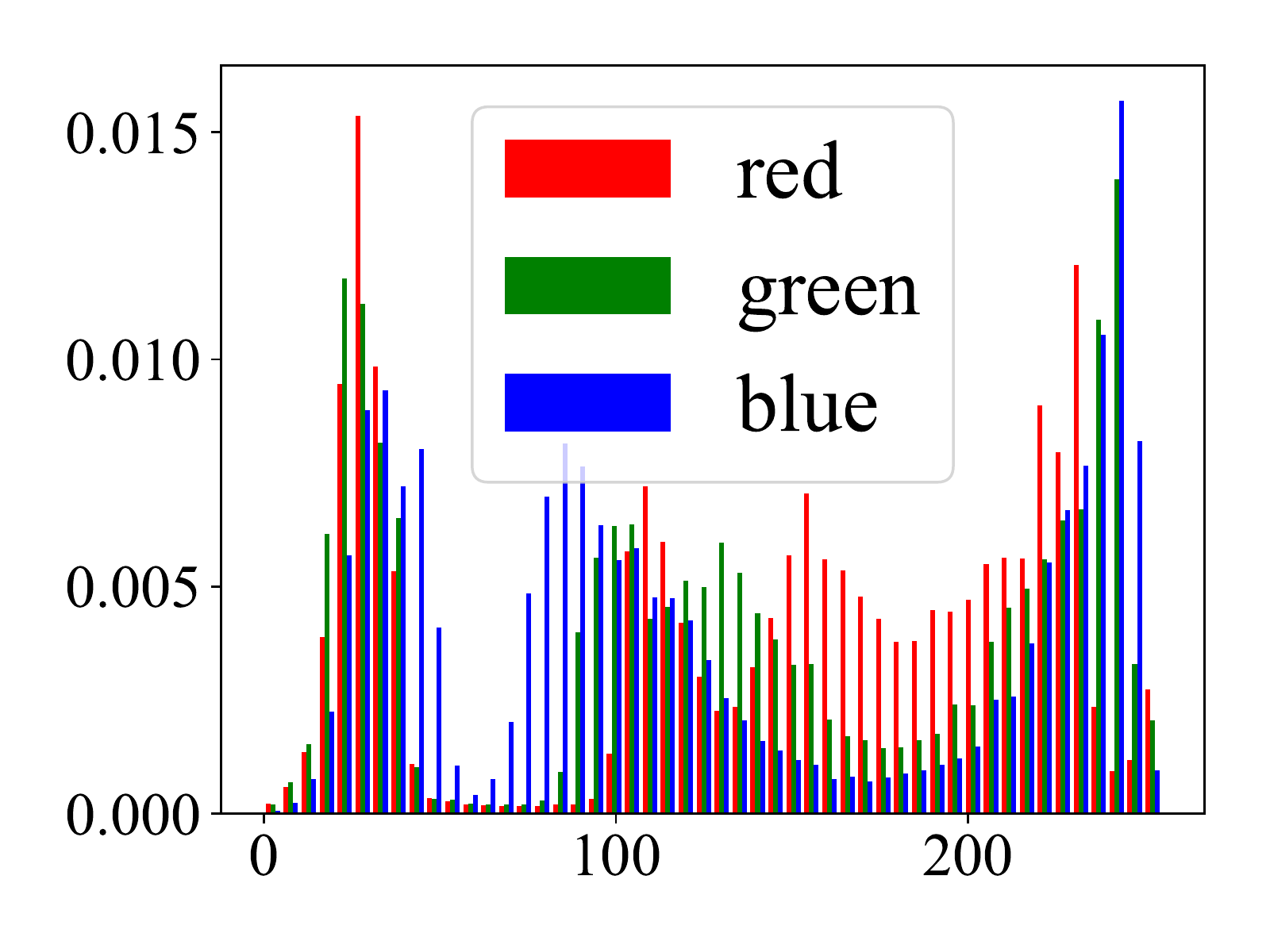}
    \label{fig:RGBhist}}
\subfloat[]{
    \includegraphics[width=0.45\linewidth]{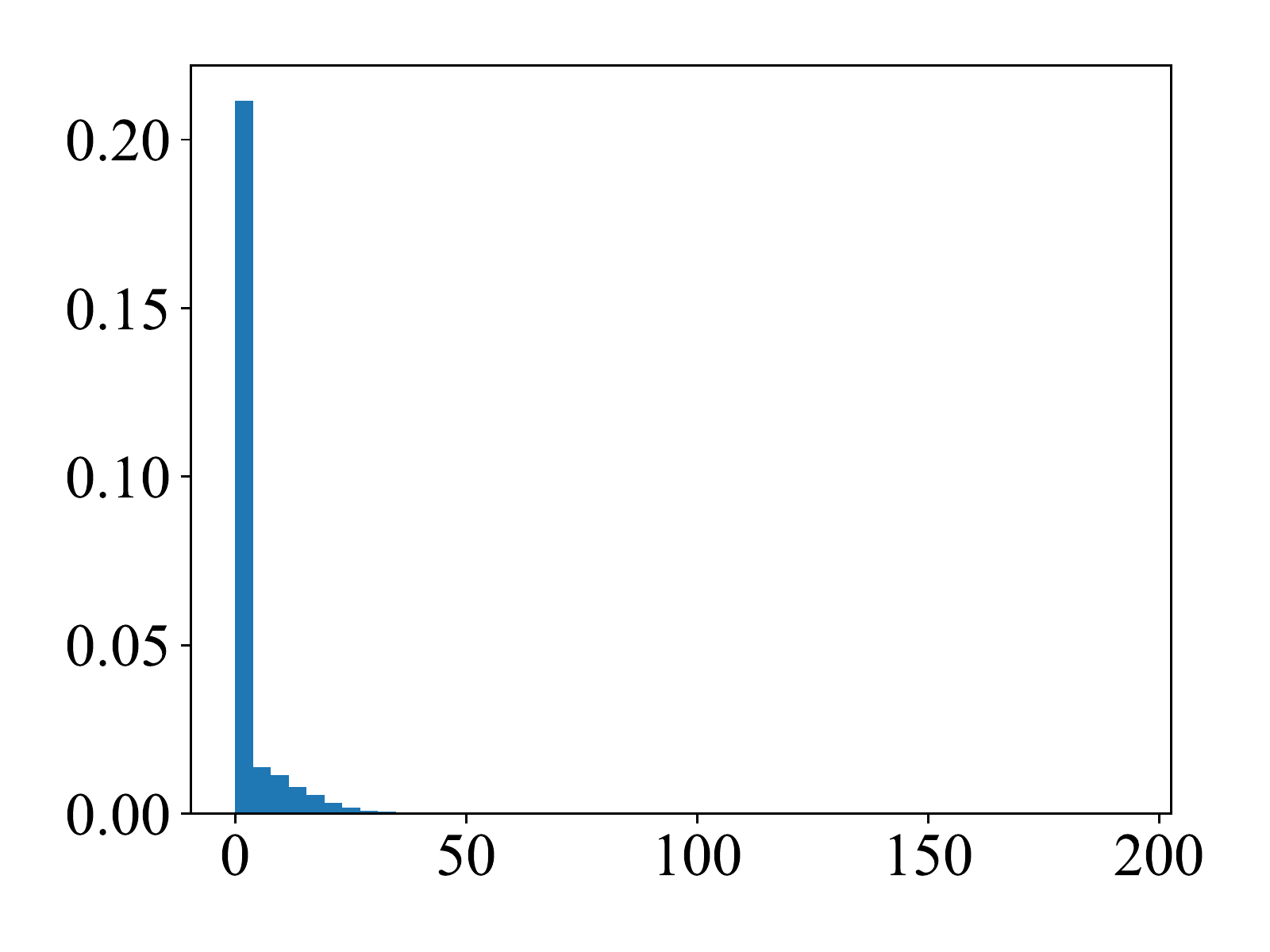}
     \label{fig:PMhist}}
\caption{Histogram comparison of (a) RGB, and (b) pressure map (PM) images shown in \figref{comparison}(d).}
\label{fig:hist}
\end{figure}
}

\newcommand{\stgsRGB}{
\begin{figure}[t]
    \centering
    \includegraphics[width=0.95\linewidth]{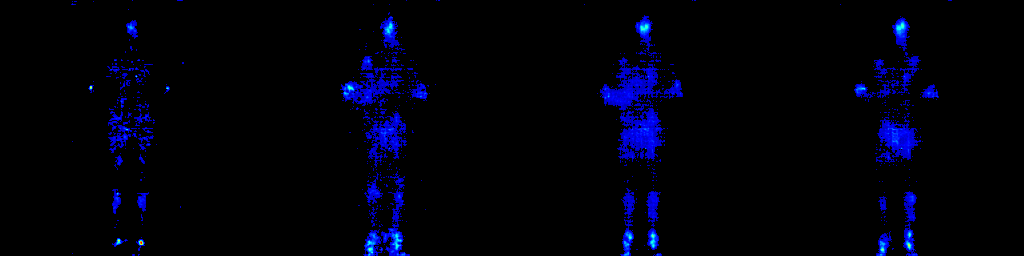}
    \caption{PM estimation outcomes of PEye network in pwrs-phy configuration with varying stages, with an RGB image as the input modality. The ground truth PM is on far left side followed by the PEye network results with 1-stage, 2-stage, and 3-stage configurations.}
    \label{fig:stgsRGB}
\end{figure}
}

\newcommand{\figGridsSht}{
\begin{figure}[t]
    \centering
    \includegraphics[width=.95\linewidth]{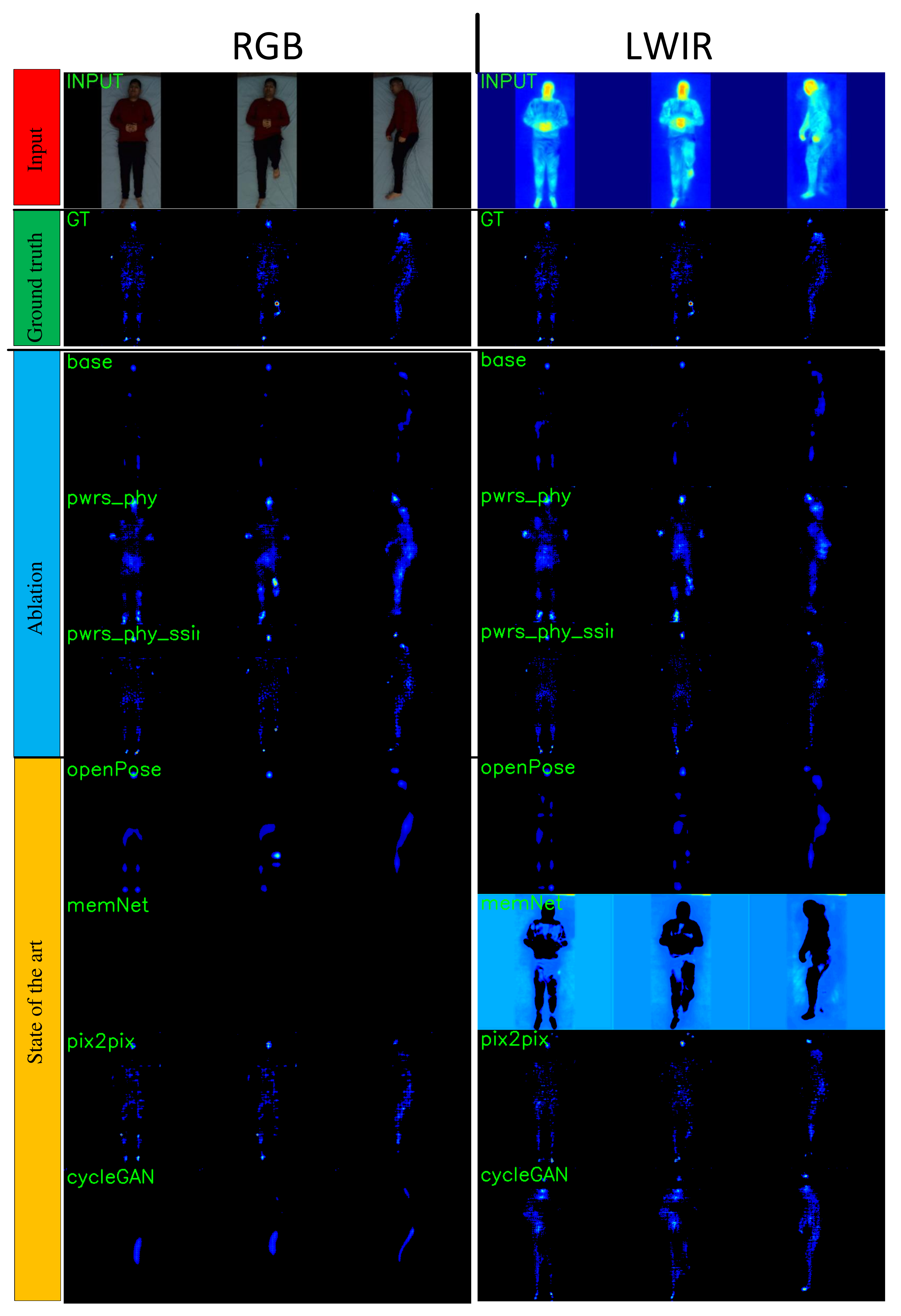}
    \caption{Qualitative results of PEye network ablation study with RGB and LWIR inputs, compared to the state-of-the-art.}
    \label{fig:gridsSht}
\end{figure}
}

\newcommand{\figAbla}{
\begin{figure}[ht]
\subfloat[]{
    \centering
    \includegraphics[width=0.5\linewidth]{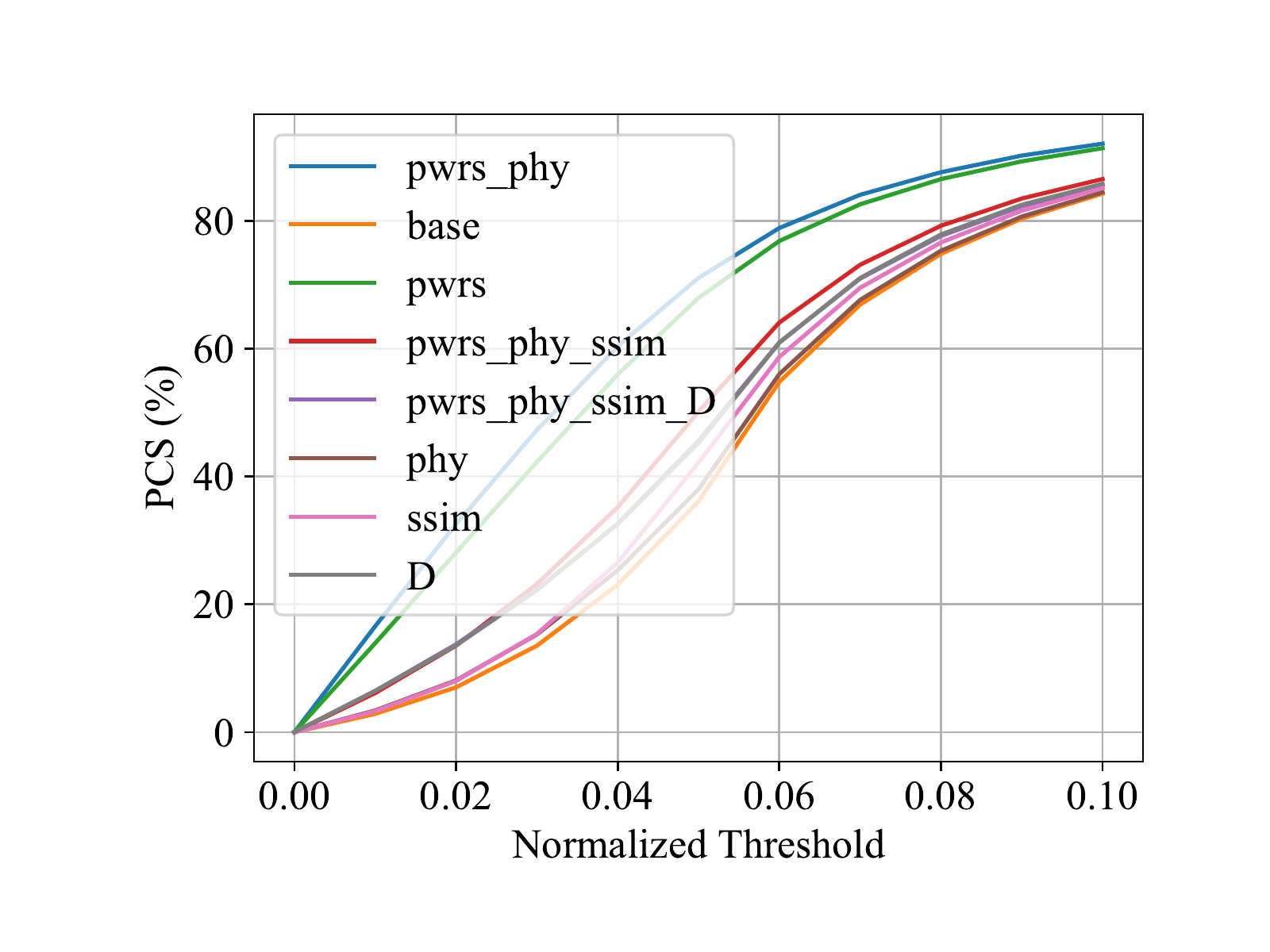}
    }
\subfloat[]{
    \centering
    \includegraphics[width=0.5\linewidth]{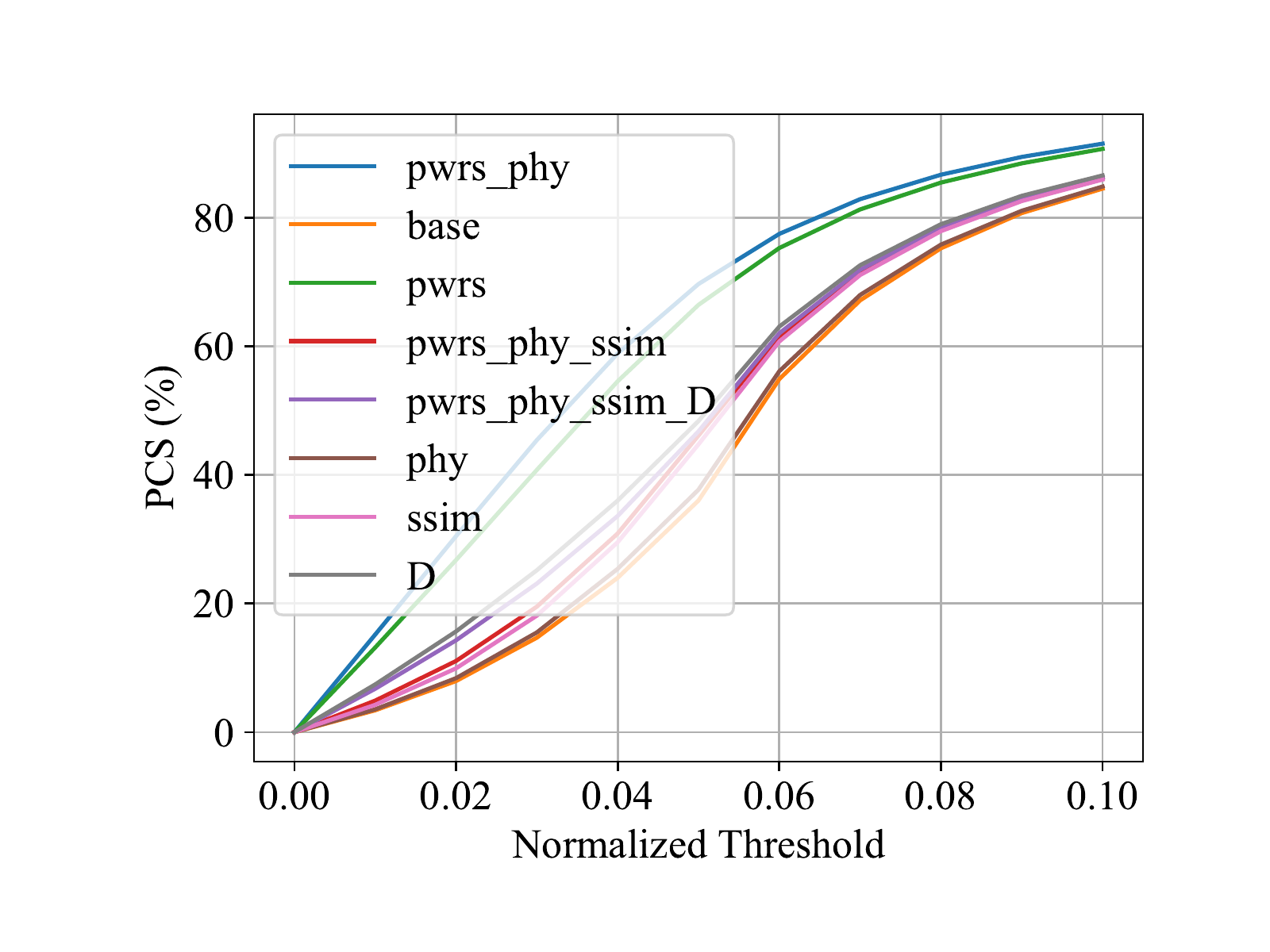}
     }
    \caption{$PCS_{efs}$ plots for PEye network ablation study with input domain as (a) RGB, and (b) LWIR.}
    \label{fig:abla}
\end{figure}
}

\newcommand{\figSota}{
\begin{figure}[ht]
\subfloat[]{
    \centering
    \includegraphics[width=0.5\linewidth]{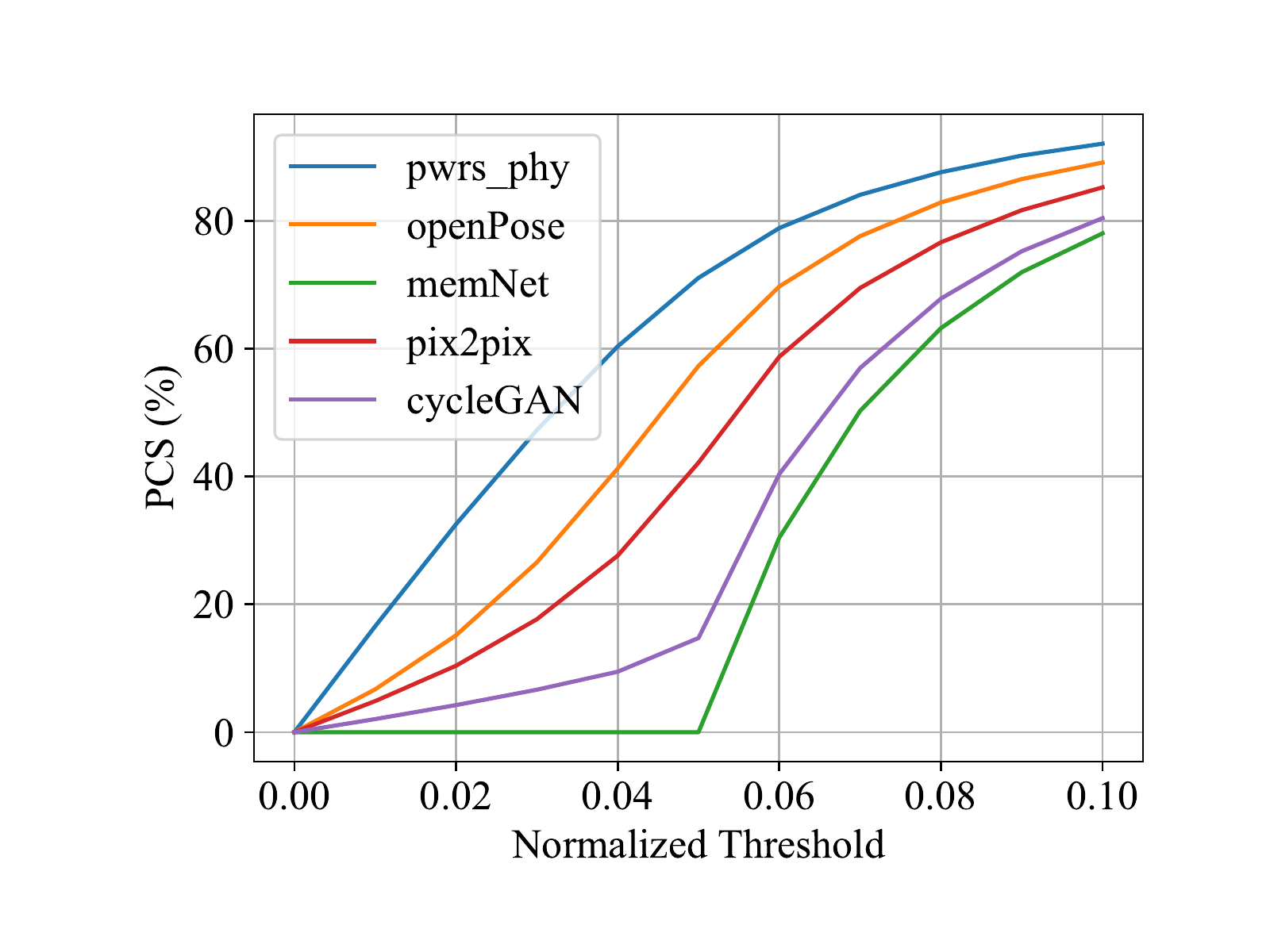}
    }
\subfloat[]{
    \centering
    \includegraphics[width=0.5\linewidth]{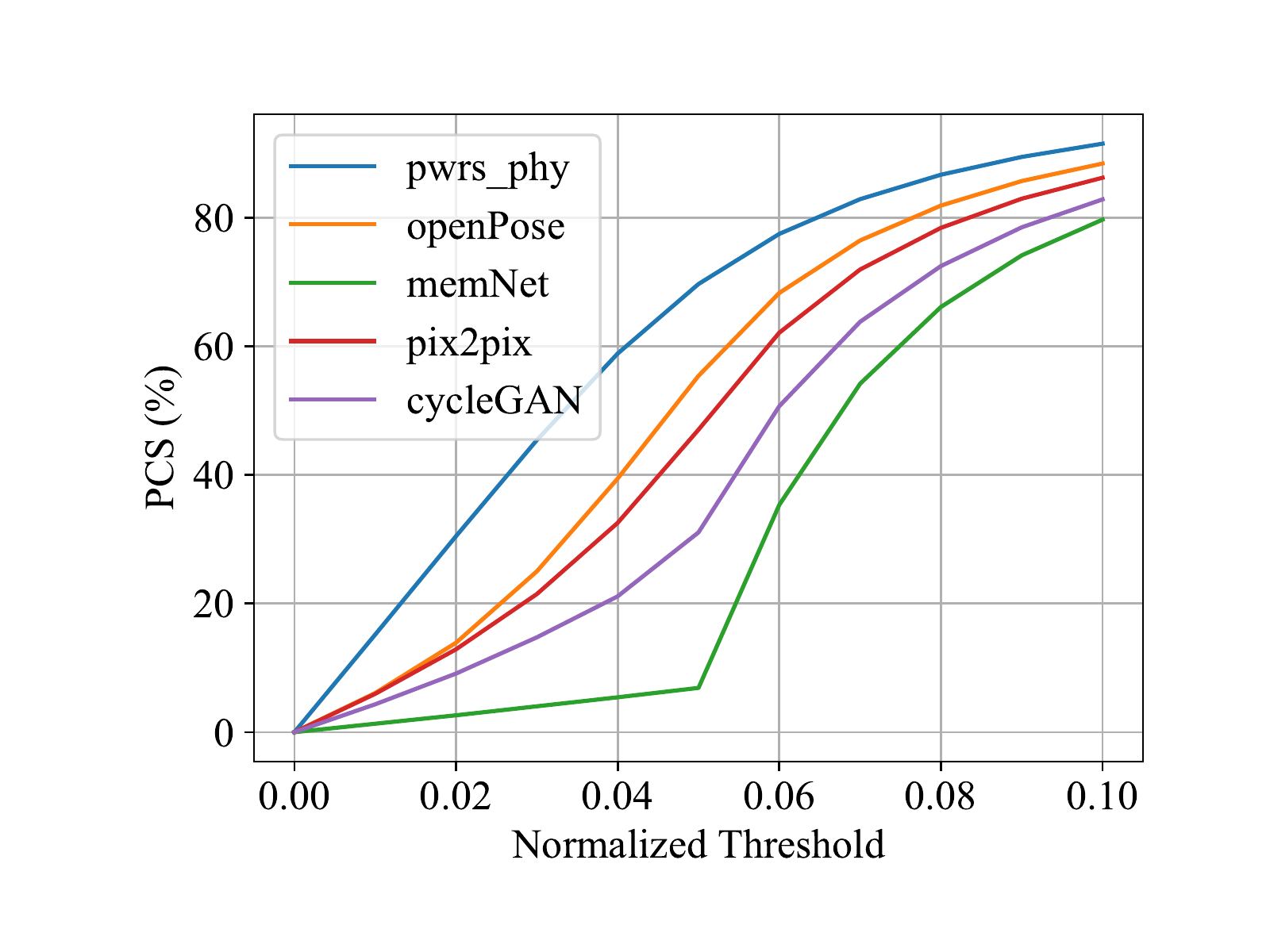}
     }
    \caption{$PCS_{efs}$ plots comparing the PEye approach with the state-of-the-art with input domain as (a) RGB, and (b) LWIR.}
    \label{fig:sota}
\end{figure}
}

\newcommand{\figPhyNum}{
\begin{figure}[t]
\subfloat[]{\centering \label{fig:nPhy_RGB}
    \includegraphics[width=0.5\linewidth]{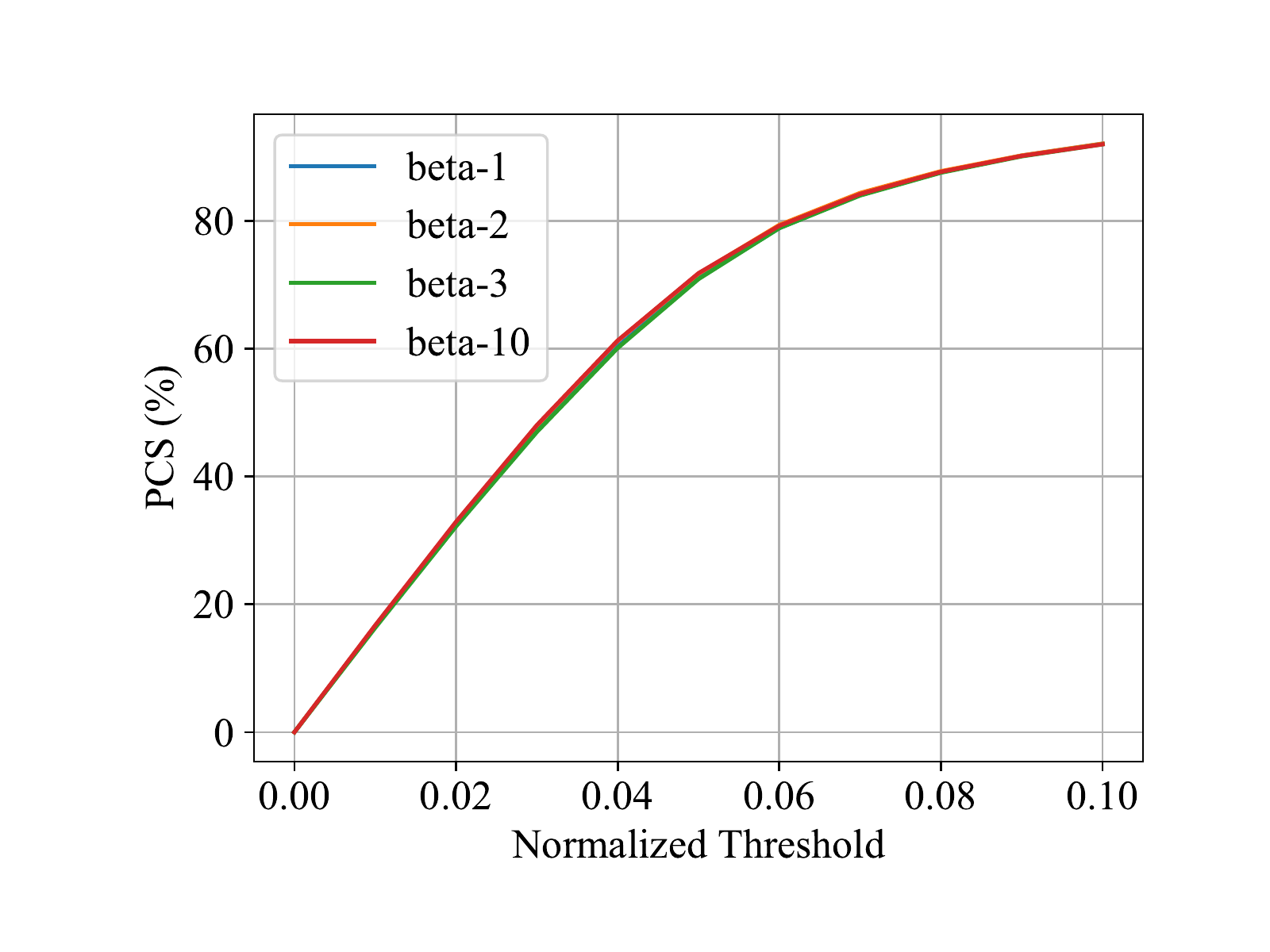}
    }
\subfloat[]{\centering \label{fig:nPhy_RGB_zoomIn}
    \includegraphics[width=0.5\linewidth]{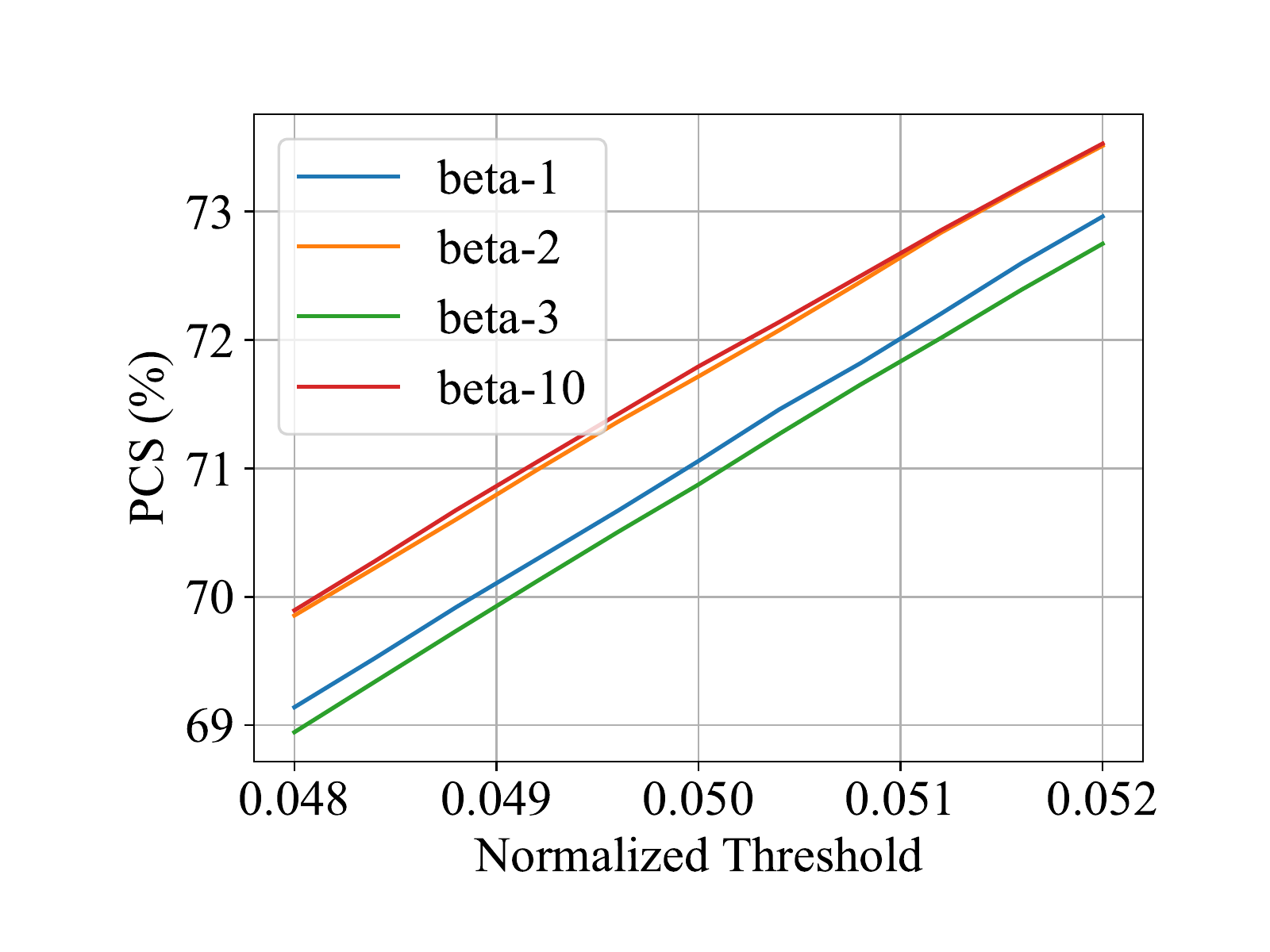}
     } \\
 \subfloat[]{\centering \label{fig:nPhy_IR}
    \includegraphics[width=0.5\linewidth]{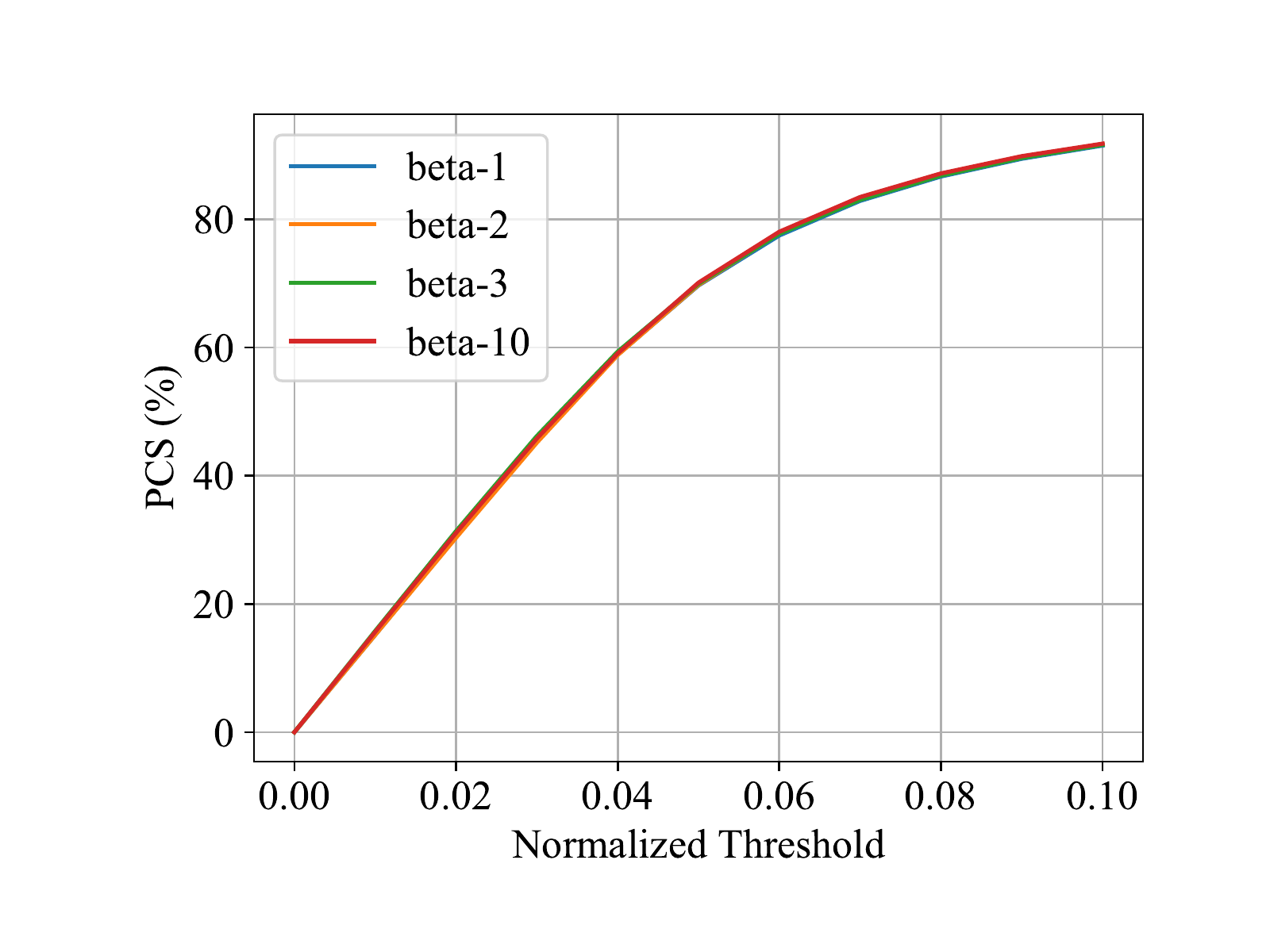}
     }
      \subfloat[]{\centering \label{fig:nPhy_IR_zoomIn}
    \includegraphics[width=0.5\linewidth]{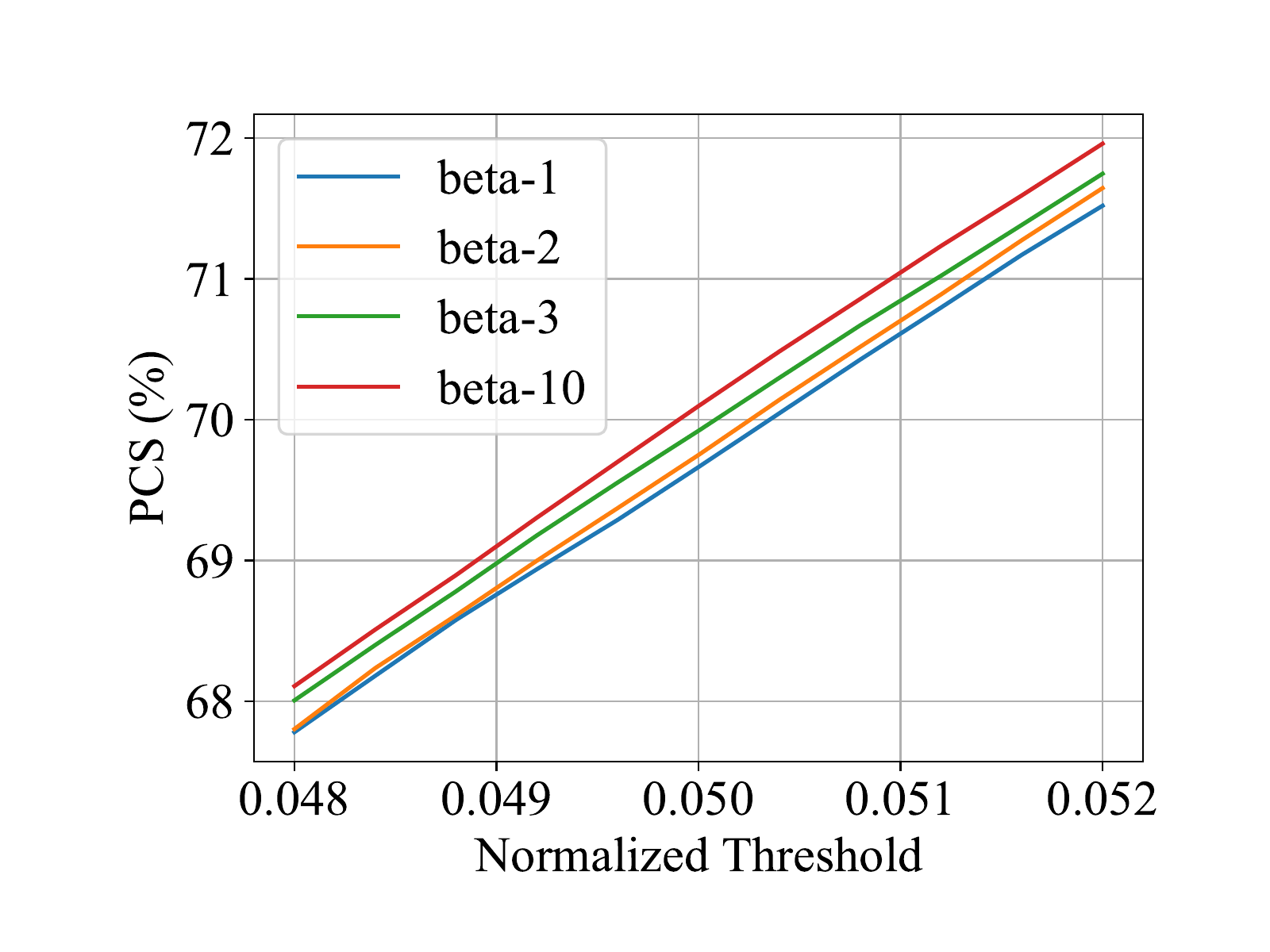}
     }
    \caption{Performance of PEye network with pwrs-phy configuration with different physical vector $\beta$ when input domain is: (A) RGB, (b) RGB zoomed-in, (c) LWIR, (d) LWIR zoomed-in.}
    \label{fig:phyNum}
\end{figure}
}

%% file: tabs.tex
\newcommand{\tblAblaRGB}{
\begin{table}
\caption{Performance of different PEye network configurations with RGB as input image modality.}
\vspace{-.2in}
\begin{center}
\footnotesize
\resizebox{\columnwidth}{!}{%
 \begin{tabular}{|c | c  | c  |  c  |   c |  c |   c  }
 \hline
Models/Metrics  & $MSE_{efs}$(e-3) & $PSNR$ & $PCS_{efs0.05}$ & $PCS_{efs0.1}$  & $SSIM$  \\
\hline
base         & 9.04          & 80.76   & 0.360         & 0.839    & 0.956     \\
pwrs           & 5.22        & 75.36   & 0.673         & 0.910     & 0.919    \\
pwrs-phy  & \cellcolor{Gray}\textbf{4.80} & 73.36  & \cellcolor{Gray}\textbf{0.707}& \cellcolor{Gray}\textbf{0.918} & 0.906\\
pwrs-phy-ssim   & 8.83      & 79.73  & 0.498         & 0.861      & 0.959           \\
pwrs-phy-ssim-D & 8.28          & 78.60 & 0.455       & 0.852     & 0.953     \\
phy          & 8.92          & 80.73  & 0.379         & 0.841    & 0.956         \\
ssim        & 8.65       & \cellcolor{Gray}\textbf{81.36}  & 0.420   & 0.848     & \cellcolor{Gray}\textbf{0.960}          \\
D           & 8.21          & 78.38     & 0.451     & 0.854     & 0.950         \\
 \hline
\end{tabular}
}
\label{tbl:ablaRGB}
\vspace{-.1in}
\end{center}
\end{table}

}

\newcommand{\tblAblaIR}{
\begin{table}
\caption{Performance of different PEye network configurations with LWIR as input image modality.}
\vspace{-.2in}
\begin{center}
\footnotesize
\resizebox{\columnwidth}{!}{%
 \begin{tabular}{|c | c  | c  |  c  |   c |  c |   c  }
 \hline
Models/Metrics  &$MSE_{efs}$(e-3) & $PSNR$  & $PCS_{efs0.05}$ & $PCS_{efs0.1}$  &$SSIM$ \\
\hline
base	        & 8.90 &	81.01 &	0.359 &	0.841 &	0.957 	\\
pwrs	        & 5.18 &	72.63 &	0.662 &	0.904 &	0.901 	\\
pwrs-phy	    & \cellcolor{Gray}\textbf{4.81} &	71.50 &	\cellcolor{Gray}\textbf{0.695} &	\cellcolor{Gray}\textbf{0.912} &	0.887 	 \\
pwrs-phy-ssim	& 8.14 &	\cellcolor{Gray}\textbf{81.74} &	0.462 &	0.859 &	\cellcolor{Gray}\textbf{0.962} 	\\
pwrs-phy-ssim-D	& 8.02 &	78.85 &	0.469 &	0.859 &	0.954 	\\
phy	            & 8.78 &	81.14 &	0.375 &	0.845 &	0.957 	\\
ssim	        & 8.29 &	81.68 &	0.447 &	0.856 &	0.962 	\\
D	            & 7.85 &	78.06 &	0.483 &	0.862 &	0.949 	\\

 \hline
\end{tabular}
}
\label{tbl:ablaIR}
\end{center}
\vspace{-.1in}
\end{table}
}

\newcommand{\tblstg}{
\begin{table}[t]
\caption{Performance of  PEye network with different numbers of stages in pwrs-phy configuration with RGB or LWIR as input image modality.}
\begin{center}
\footnotesize
\resizebox{\columnwidth}{!}{%
 \begin{tabular}{|c | c  | c  |  c  |   c |  c |   c  }
 \hline
RGB  & $MSE_{efs}$(e-3) & $PSNR$  & $PCS_{efs0.05}$ & $PCS_{efs0.1}$  &$SSIM$\\
\hline      
1-Stage &	4.95 &	72.41 &	0.692 &	0.915 &	0.839 	\\ 
2-Stage &	4.78 &	73.12 &	0.722 &	0.918 &	0.894 	\\ 
3-Stage &	4.80 &	73.41 &	0.707 &	0.918 &	0.906 \\ 
\hline
\hline
LWIR  & $MSE_{efs}$(e-3) & $PSNR$  & $PCS_{efs0.05}$ & $PCS_{efs0.1}$  &$SSIM$\\
\hline      
1-Stage &	4.82 	&69.44 	&0.689 	&0.910 	&0.834 	\\ 
2-Stage &	4.76 	&70.91 	&0.699 	&0.913 	&0.876 	\\ 
3-Stage &	4.81 	&71.50 	&0.695 	&0.912 	&0.887 	\\ 

\hline
\end{tabular}
}
\end{center}
\label{tbl:stg}
\vspace{-.1in}
\end{table}
}

\newcommand{\tblSotaRGB}{
\begin{table}[t]
\caption{Performance comparison of PEye network with pwrs-phy configuration with state-of-the-art, with RGB as input image modality.}
\vspace{-.2in}
\begin{center}
\footnotesize
\resizebox{\columnwidth}{!}{%
 \begin{tabular}{|c | c  | c  |  c  |   c |  c |   c  }
 \hline
Models/Metrics  & $MSE_{efs}$(e-3) & $PSNR$ & $PCS_{efs0.05}$ & $PCS_{efs0.1}$ & $SSIM$   \\
\hline
pwrs-phy &	\cellcolor{Gray}\textbf{4.80} &	73.41 &	\cellcolor{Gray}\textbf{0.707} &	\cellcolor{Gray}\textbf{0.918} &	0.906 	\\ 
openPose &	6.49 &	\cellcolor{Gray}\textbf{82.29} &	0.568 &	0.888 &	\cellcolor{Gray}\textbf{0.958} 	\\ 
memNet &	11.56 &	78.49 &	0.000 &	0.779 &	0.941 	\\ 
pix2pix &	33.64 &	73.15 &	0.423 &	0.849 &	0.952 	\\ 
cycleGAN &	42.28 &	71.49 &	0.148 &	0.803 &	0.951 	\\ 
 \hline
\end{tabular}
}
\label{tbl:sotaRGB}
\end{center}
\vspace{-.1in}
\end{table}
}

\newcommand{\tblSotaIR}{
\begin{table}[t]
\caption{Performance comparison of PEye network with pwrs-phy configuration with state-of-the-art, with LWIR as input image modality.}
\vspace{-.2in}
\begin{center}
\footnotesize
\resizebox{\columnwidth}{!}{%
 \begin{tabular}{|c | c  | c  |  c  |   c |  c |   c  }
 \hline
Models/Metrics  & $MSE_{efs}$(e-3) & $PSNR$  & $PCS_{efs0.05}$ & $PCS_{efs0.1}$  &$SSIM$\\
\hline
pwrs-phy &	\cellcolor{Gray}\textbf{4.81} &	71.50 &	\cellcolor{Gray}\textbf{0.695} &	\cellcolor{Gray}\textbf{0.912} &	0.887 	\\ 
openPose &	6.98 &	\cellcolor{Gray}\textbf{82.10} &	0.551 &	0.881 &	\cellcolor{Gray}\textbf{0.957} 	\\ 
memNet &	10.92 &	36.08 &	0.067 &	0.795 &	0.089 	\\ 
pix2pix &	8.07 &	78.74 &	0.469 &	0.858 &	0.950 	\\ 
cycleGAN &	9.35 &	74.32 &	0.309 &	0.826 &	0.910 	\\ 

\hline
\end{tabular}
}
\label{tbl:sotaIR}
\end{center}
\vspace{-.1in}
\end{table}
}

\newcommand{\tblBetaB}{
\begin{table*}[ht]
\caption{Physical parameter for PADS-PM listed in $\beta$.}
\begin{center}
{\small
\begin{tabular}{ |c |c| c| c| c |}
\hline
\multicolumn{5}{|c|}{List of Physical Parameters in vector $\beta$} \\
\hline 
    weight (kg) & height (cm) & gender [0|1] & bust (cm) & waist (cm)  \\ 
\hline
    hip (cm) & upperArm-R (cm) & lowerArm-R (cm) & thigh-R (cm) & shank-R (cm) \\
\hline
\end{tabular}
}
\end{center}
\label{tbl:betaB}
\end{table*}
}


%% file: paper.bbl
\begin{thebibliography}{47}
\expandafter\ifx\csname natexlab\endcsname\relax\def\natexlab#1{#1}\fi
\providecommand{\url}[1]{\texttt{#1}}
\providecommand{\href}[2]{#2}
\providecommand{\path}[1]{#1}
\providecommand{\DOIprefix}{doi:}
\providecommand{\ArXivprefix}{arXiv:}
\providecommand{\URLprefix}{URL: }
\providecommand{\Pubmedprefix}{pmid:}
\providecommand{\doi}[1]{\href{http://dx.doi.org/#1}{\path{#1}}}
\providecommand{\Pubmed}[1]{\href{pmid:#1}{\path{#1}}}
\providecommand{\bibinfo}[2]{#2}
\ifx\xfnm\relax \def\xfnm[#1]{\unskip,\space#1}\fi
\bibitem[{Andriluka et~al.(2014)Andriluka, Pishchulin, Gehler and
  Schiele}]{andriluka20142d}
\bibinfo{author}{Andriluka, M.}, \bibinfo{author}{Pishchulin, L.},
  \bibinfo{author}{Gehler, P.}, \bibinfo{author}{Schiele, B.},
  \bibinfo{year}{2014}.
\newblock \bibinfo{title}{2d human pose estimation: New benchmark and state of
  the art analysis}, in: \bibinfo{booktitle}{Proceedings of the IEEE Conference
  on computer Vision and Pattern Recognition}, pp. \bibinfo{pages}{3686--3693}.
\bibitem[{Batista et~al.(2004)Batista, Prati and Monard}]{batista2004study}
\bibinfo{author}{Batista, G.E.}, \bibinfo{author}{Prati, R.C.},
  \bibinfo{author}{Monard, M.C.}, \bibinfo{year}{2004}.
\newblock \bibinfo{title}{A study of the behavior of several methods for
  balancing machine learning training data}.
\newblock \bibinfo{journal}{ACM SIGKDD explorations newsletter}
  \bibinfo{volume}{6}, \bibinfo{pages}{20--29}.
\bibitem[{Bishop(2006)}]{bishop2006pattern}
\bibinfo{author}{Bishop, C.M.}, \bibinfo{year}{2006}.
\newblock \bibinfo{title}{Pattern recognition and machine learning}.
\newblock \bibinfo{publisher}{springer}.
\bibitem[{Black et~al.(2007)Black, Baharestani, Cuddigan, Dorner, Edsberg,
  Langemo, Posthauer, Ratliff, Taler et~al.}]{black2007national}
\bibinfo{author}{Black, J.}, \bibinfo{author}{Baharestani, M.M.},
  \bibinfo{author}{Cuddigan, J.}, \bibinfo{author}{Dorner, B.},
  \bibinfo{author}{Edsberg, L.}, \bibinfo{author}{Langemo, D.},
  \bibinfo{author}{Posthauer, M.E.}, \bibinfo{author}{Ratliff, C.},
  \bibinfo{author}{Taler, G.}, et~al., \bibinfo{year}{2007}.
\newblock \bibinfo{title}{National pressure ulcer advisory panel's updated
  pressure ulcer staging system}.
\newblock \bibinfo{journal}{Advances in skin \& wound care}
  \bibinfo{volume}{20}, \bibinfo{pages}{269--274}.
\bibitem[{Brock et~al.(2018)Brock, Donahue and Simonyan}]{brock2018large}
\bibinfo{author}{Brock, A.}, \bibinfo{author}{Donahue, J.},
  \bibinfo{author}{Simonyan, K.}, \bibinfo{year}{2018}.
\newblock \bibinfo{title}{Large scale gan training for high fidelity natural
  image synthesis}.
\newblock \bibinfo{journal}{arXiv preprint arXiv:1809.11096} .
\bibitem[{Cao et~al.(2018)Cao, Hidalgo, Simon, Wei and
  Sheikh}]{cao2018openpose}
\bibinfo{author}{Cao, Z.}, \bibinfo{author}{Hidalgo, G.},
  \bibinfo{author}{Simon, T.}, \bibinfo{author}{Wei, S.E.},
  \bibinfo{author}{Sheikh, Y.}, \bibinfo{year}{2018}.
\newblock \bibinfo{title}{Open{P}ose: realtime multi-person 2{D} pose
  estimation using {P}art {A}ffinity {F}ields}, in: \bibinfo{booktitle}{arXiv
  preprint arXiv:1812.08008}.
\bibitem[{Chawla et~al.(2002)Chawla, Bowyer, Hall and
  Kegelmeyer}]{chawla2002smote}
\bibinfo{author}{Chawla, N.V.}, \bibinfo{author}{Bowyer, K.W.},
  \bibinfo{author}{Hall, L.O.}, \bibinfo{author}{Kegelmeyer, W.P.},
  \bibinfo{year}{2002}.
\newblock \bibinfo{title}{Smote: synthetic minority over-sampling technique}.
\newblock \bibinfo{journal}{Journal of artificial intelligence research}
  \bibinfo{volume}{16}, \bibinfo{pages}{321--357}.
\bibitem[{Clever et~al.(2018)Clever, Kapusta, Park, Erickson, Chitalia and
  Kemp}]{clever20183d}
\bibinfo{author}{Clever, H.M.}, \bibinfo{author}{Kapusta, A.},
  \bibinfo{author}{Park, D.}, \bibinfo{author}{Erickson, Z.},
  \bibinfo{author}{Chitalia, Y.}, \bibinfo{author}{Kemp, C.C.},
  \bibinfo{year}{2018}.
\newblock \bibinfo{title}{3d human pose estimation on a configurable bed from a
  pressure image}, in: \bibinfo{booktitle}{2018 IEEE/RSJ International
  Conference on Intelligent Robots and Systems (IROS)},
  \bibinfo{organization}{IEEE}. pp. \bibinfo{pages}{54--61}.
\bibitem[{Febriana et~al.(2019)Febriana, Rizal and Susanto}]{febriana2019sleep}
\bibinfo{author}{Febriana, N.}, \bibinfo{author}{Rizal, A.},
  \bibinfo{author}{Susanto, E.}, \bibinfo{year}{2019}.
\newblock \bibinfo{title}{Sleep monitoring system based on body posture
  movement using microsoft kinect sensor}, in: \bibinfo{booktitle}{AIP
  Conference Proceedings}, \bibinfo{organization}{AIP Publishing LLC}. p.
  \bibinfo{pages}{020012}.
\bibitem[{Greminger and Nelson(2004)}]{greminger2004vision}
\bibinfo{author}{Greminger, M.A.}, \bibinfo{author}{Nelson, B.J.},
  \bibinfo{year}{2004}.
\newblock \bibinfo{title}{Vision-based force measurement}.
\newblock \bibinfo{journal}{IEEE Transactions on Pattern Analysis and Machine
  Intelligence} \bibinfo{volume}{26}, \bibinfo{pages}{290--298}.
\bibitem[{Hartley and Zisserman(2003)}]{hartley2003multiple}
\bibinfo{author}{Hartley, R.}, \bibinfo{author}{Zisserman, A.},
  \bibinfo{year}{2003}.
\newblock \bibinfo{title}{Multiple view geometry in computer vision}.
\newblock \bibinfo{publisher}{Cambridge university press}.
\bibitem[{Herland et~al.(2018)Herland, Khoshgoftaar and
  Bauder}]{herland2018big}
\bibinfo{author}{Herland, M.}, \bibinfo{author}{Khoshgoftaar, T.M.},
  \bibinfo{author}{Bauder, R.A.}, \bibinfo{year}{2018}.
\newblock \bibinfo{title}{Big data fraud detection using multiple medicare data
  sources}.
\newblock \bibinfo{journal}{Journal of Big Data} \bibinfo{volume}{5},
  \bibinfo{pages}{29}.
\bibitem[{Isola et~al.(2017)Isola, Zhu, Zhou and Efros}]{isola2017image}
\bibinfo{author}{Isola, P.}, \bibinfo{author}{Zhu, J.Y.},
  \bibinfo{author}{Zhou, T.}, \bibinfo{author}{Efros, A.A.},
  \bibinfo{year}{2017}.
\newblock \bibinfo{title}{Image-to-image translation with conditional
  adversarial networks}, in: \bibinfo{booktitle}{Proceedings of the IEEE
  conference on computer vision and pattern recognition}, pp.
  \bibinfo{pages}{1125--1134}.
\bibitem[{Japkowicz(2000)}]{japkowicz2000class}
\bibinfo{author}{Japkowicz, N.}, \bibinfo{year}{2000}.
\newblock \bibinfo{title}{The class imbalance problem: Significance and
  strategies}, in: \bibinfo{booktitle}{Proc. of the Int’l Conf. on Artificial
  Intelligence}, \bibinfo{organization}{Citeseer}.
\bibitem[{Johnson and Khoshgoftaar(2019)}]{johnson2019survey}
\bibinfo{author}{Johnson, J.M.}, \bibinfo{author}{Khoshgoftaar, T.M.},
  \bibinfo{year}{2019}.
\newblock \bibinfo{title}{Survey on deep learning with class imbalance}.
\newblock \bibinfo{journal}{Journal of Big Data} \bibinfo{volume}{6},
  \bibinfo{pages}{27}.
\bibitem[{Kingma and Ba(2014)}]{kingma2014adam}
\bibinfo{author}{Kingma, D.P.}, \bibinfo{author}{Ba, J.}, \bibinfo{year}{2014}.
\newblock \bibinfo{title}{Adam: A method for stochastic optimization}.
\newblock \bibinfo{journal}{arXiv preprint arXiv:1412.6980} .
\bibitem[{Kubat et~al.(1997)Kubat, Matwin et~al.}]{kubat1997addressing}
\bibinfo{author}{Kubat, M.}, \bibinfo{author}{Matwin, S.}, et~al.,
  \bibinfo{year}{1997}.
\newblock \bibinfo{title}{Addressing the curse of imbalanced training sets:
  one-sided selection}, in: \bibinfo{booktitle}{Icml},
  \bibinfo{organization}{Citeseer}. pp. \bibinfo{pages}{179--186}.
\bibitem[{Ledig et~al.(2017)Ledig, Theis, Husz{\'a}r, Caballero, Cunningham,
  Acosta, Aitken, Tejani, Totz, Wang et~al.}]{ledig2017photo}
\bibinfo{author}{Ledig, C.}, \bibinfo{author}{Theis, L.},
  \bibinfo{author}{Husz{\'a}r, F.}, \bibinfo{author}{Caballero, J.},
  \bibinfo{author}{Cunningham, A.}, \bibinfo{author}{Acosta, A.},
  \bibinfo{author}{Aitken, A.}, \bibinfo{author}{Tejani, A.},
  \bibinfo{author}{Totz, J.}, \bibinfo{author}{Wang, Z.}, et~al.,
  \bibinfo{year}{2017}.
\newblock \bibinfo{title}{Photo-realistic single image super-resolution using a
  generative adversarial network}, in: \bibinfo{booktitle}{Proceedings of the
  IEEE conference on computer vision and pattern recognition}, pp.
  \bibinfo{pages}{4681--4690}.
\bibitem[{Liu et~al.(2014)Liu, Huang, Xu, Zhang, Stevens, Alshurafa and
  Sarrafzadeh}]{liu2014breathsens}
\bibinfo{author}{Liu, J.J.}, \bibinfo{author}{Huang, M.C.},
  \bibinfo{author}{Xu, W.}, \bibinfo{author}{Zhang, X.},
  \bibinfo{author}{Stevens, L.}, \bibinfo{author}{Alshurafa, N.},
  \bibinfo{author}{Sarrafzadeh, M.}, \bibinfo{year}{2014}.
\newblock \bibinfo{title}{Breathsens: A continuous on-bed respiratory
  monitoring system with torso localization using an unobtrusive pressure
  sensing array}.
\newblock \bibinfo{journal}{IEEE journal of biomedical and health informatics}
  \bibinfo{volume}{19}, \bibinfo{pages}{1682--1688}.
\bibitem[{Liu et~al.(2020)Liu, Huang, Fu, Li, Su and
  Ostadabbas}]{liu2020simultaneously}
\bibinfo{author}{Liu, S.}, \bibinfo{author}{Huang, X.}, \bibinfo{author}{Fu,
  N.}, \bibinfo{author}{Li, C.}, \bibinfo{author}{Su, Z.},
  \bibinfo{author}{Ostadabbas, S.}, \bibinfo{year}{2020}.
\newblock \bibinfo{title}{Simultaneously-collected multimodal lying pose
  dataset: Towards in-bed human pose monitoring under adverse vision
  conditions}.
\newblock \bibinfo{journal}{arXiv preprint arXiv:2008.08735} .
\bibitem[{Liu and Ostadabbas(2017)}]{liu2017vision}
\bibinfo{author}{Liu, S.}, \bibinfo{author}{Ostadabbas, S.},
  \bibinfo{year}{2017}.
\newblock \bibinfo{title}{A vision-based system for in-bed posture tracking},
  in: \bibinfo{booktitle}{Proceedings of the IEEE International Conference on
  Computer Vision}, pp. \bibinfo{pages}{1373--1382}.
\bibitem[{Liu and Ostadabbas(2019)}]{liu2019seeing}
\bibinfo{author}{Liu, S.}, \bibinfo{author}{Ostadabbas, S.},
  \bibinfo{year}{2019}.
\newblock \bibinfo{title}{Seeing under the cover: A physics guided learning
  approach for in-bed pose estimation}, in: \bibinfo{booktitle}{International
  Conference on Medical Image Computing and Computer-Assisted Intervention},
  \bibinfo{organization}{Springer}. pp. \bibinfo{pages}{236--245}.
\bibitem[{Liu et~al.(2019)Liu, Yin and Ostadabbas}]{liu2019bed}
\bibinfo{author}{Liu, S.}, \bibinfo{author}{Yin, Y.},
  \bibinfo{author}{Ostadabbas, S.}, \bibinfo{year}{2019}.
\newblock \bibinfo{title}{In-bed pose estimation: Deep learning with shallow
  dataset}.
\newblock \bibinfo{journal}{IEEE journal of translational engineering in health
  and medicine} \bibinfo{volume}{7}, \bibinfo{pages}{1--12}.
\bibitem[{Long et~al.(2015)Long, Shelhamer and Darrell}]{long2015fully}
\bibinfo{author}{Long, J.}, \bibinfo{author}{Shelhamer, E.},
  \bibinfo{author}{Darrell, T.}, \bibinfo{year}{2015}.
\newblock \bibinfo{title}{Fully convolutional networks for semantic
  segmentation}.
\newblock \bibinfo{journal}{Proceedings of the IEEE conference on Computer
  Vision and Pattern Recognition} , \bibinfo{pages}{3431--3440}.
\bibitem[{Martinez et~al.(2015)Martinez, Rybok and
  Stiefelhagen}]{martinez2015action}
\bibinfo{author}{Martinez, M.}, \bibinfo{author}{Rybok, L.},
  \bibinfo{author}{Stiefelhagen, R.}, \bibinfo{year}{2015}.
\newblock \bibinfo{title}{Action recognition in bed using bams for assisted
  living and elderly care}, in: \bibinfo{booktitle}{2015 14th IAPR
  International Conference on Machine Vision Applications (MVA)},
  \bibinfo{organization}{IEEE}. pp. \bibinfo{pages}{329--332}.
\bibitem[{Mendon{\c{c}}a et~al.(2019)Mendon{\c{c}}a, Mostafa, Morgado-Dias,
  Ravelo-Garcia and Penzel}]{mendoncca2019review}
\bibinfo{author}{Mendon{\c{c}}a, F.}, \bibinfo{author}{Mostafa, S.S.},
  \bibinfo{author}{Morgado-Dias, F.}, \bibinfo{author}{Ravelo-Garcia, A.G.},
  \bibinfo{author}{Penzel, T.}, \bibinfo{year}{2019}.
\newblock \bibinfo{title}{A review of approaches for sleep quality analysis}.
\newblock \bibinfo{journal}{Ieee Access} \bibinfo{volume}{7},
  \bibinfo{pages}{24527--24546}.
\bibitem[{Murthy et~al.(2009)Murthy, Van~Jaarsveld, Fei, Pavlidis,
  Harrykissoon, Lucke, Faiz and Castriotta}]{murthy2009thermal}
\bibinfo{author}{Murthy, J.N.}, \bibinfo{author}{Van~Jaarsveld, J.},
  \bibinfo{author}{Fei, J.}, \bibinfo{author}{Pavlidis, I.},
  \bibinfo{author}{Harrykissoon, R.I.}, \bibinfo{author}{Lucke, J.F.},
  \bibinfo{author}{Faiz, S.}, \bibinfo{author}{Castriotta, R.J.},
  \bibinfo{year}{2009}.
\newblock \bibinfo{title}{Thermal infrared imaging: a novel method to monitor
  airflow during polysomnography}.
\newblock \bibinfo{journal}{Sleep} \bibinfo{volume}{32},
  \bibinfo{pages}{1521--1527}.
\bibitem[{Newell et~al.(2016)Newell, Yang and Deng}]{newell2016stacked}
\bibinfo{author}{Newell, A.}, \bibinfo{author}{Yang, K.},
  \bibinfo{author}{Deng, J.}, \bibinfo{year}{2016}.
\newblock \bibinfo{title}{Stacked hourglass networks for human pose
  estimation}.
\newblock \bibinfo{journal}{European Conference on Computer Vision} ,
  \bibinfo{pages}{483--499}.
\bibitem[{Nguyen et~al.(2010)Nguyen, Cohen, Lipman, Brown, Molinari, Jackson,
  Kirking, Szymanowski, Wilson, Salhi et~al.}]{nguyen2010comparison}
\bibinfo{author}{Nguyen, A.V.}, \bibinfo{author}{Cohen, N.J.},
  \bibinfo{author}{Lipman, H.}, \bibinfo{author}{Brown, C.M.},
  \bibinfo{author}{Molinari, N.A.}, \bibinfo{author}{Jackson, W.L.},
  \bibinfo{author}{Kirking, H.}, \bibinfo{author}{Szymanowski, P.},
  \bibinfo{author}{Wilson, T.W.}, \bibinfo{author}{Salhi, B.A.}, et~al.,
  \bibinfo{year}{2010}.
\newblock \bibinfo{title}{Comparison of 3 infrared thermal detection systems
  and self-report for mass fever screening}.
\newblock \bibinfo{journal}{Emerging infectious diseases} \bibinfo{volume}{16},
  \bibinfo{pages}{1710}.
\bibitem[{Ostadabbas et~al.(2014)Ostadabbas, Pouyan, Nourani and
  Kehtarnavaz}]{ostadabbas2014bed}
\bibinfo{author}{Ostadabbas, S.}, \bibinfo{author}{Pouyan, M.B.},
  \bibinfo{author}{Nourani, M.}, \bibinfo{author}{Kehtarnavaz, N.},
  \bibinfo{year}{2014}.
\newblock \bibinfo{title}{In-bed posture classification and limb
  identification}, in: \bibinfo{booktitle}{2014 IEEE Biomedical Circuits and
  Systems Conference (BioCAS) Proceedings}, \bibinfo{organization}{IEEE}. pp.
  \bibinfo{pages}{133--136}.
\bibitem[{Ostadabbas et~al.(2015)Ostadabbas, Sebkhi, Zhang, Rahim, Anderson,
  Lee and Ghovanloo}]{ostadabbas2015vision}
\bibinfo{author}{Ostadabbas, S.}, \bibinfo{author}{Sebkhi, N.},
  \bibinfo{author}{Zhang, M.}, \bibinfo{author}{Rahim, S.},
  \bibinfo{author}{Anderson, L.J.}, \bibinfo{author}{Lee, F.E.H.},
  \bibinfo{author}{Ghovanloo, M.}, \bibinfo{year}{2015}.
\newblock \bibinfo{title}{A vision-based respiration monitoring system for
  passive airway resistance estimation}.
\newblock \bibinfo{journal}{IEEE Transactions on biomedical engineering}
  \bibinfo{volume}{63}, \bibinfo{pages}{1904--1913}.
\bibitem[{Ostadabbas et~al.(2011)Ostadabbas, Yousefi, Faezipour, Nourani and
  Pompeo}]{ostadabbas2011pressure}
\bibinfo{author}{Ostadabbas, S.}, \bibinfo{author}{Yousefi, R.},
  \bibinfo{author}{Faezipour, M.}, \bibinfo{author}{Nourani, M.},
  \bibinfo{author}{Pompeo, M.}, \bibinfo{year}{2011}.
\newblock \bibinfo{title}{Pressure ulcer prevention: An efficient turning
  schedule for bed-bound patients}.
\newblock \bibinfo{journal}{Life Science Systems and Applications Workshop
  (LiSSA), 2011 IEEE/NIH} , \bibinfo{pages}{159--162}.
\bibitem[{Ostadabbas et~al.(2012)Ostadabbas, Yousefi, Nourani, Faezipour, Tamil
  and Pompeo}]{ostadabbas2012resource}
\bibinfo{author}{Ostadabbas, S.}, \bibinfo{author}{Yousefi, R.},
  \bibinfo{author}{Nourani, M.}, \bibinfo{author}{Faezipour, M.},
  \bibinfo{author}{Tamil, L.}, \bibinfo{author}{Pompeo, M.Q.},
  \bibinfo{year}{2012}.
\newblock \bibinfo{title}{A resource-efficient planning for pressure ulcer
  prevention}.
\newblock \bibinfo{journal}{IEEE Transactions on Information Technology in
  Biomedicine} \bibinfo{volume}{16}, \bibinfo{pages}{1265--1273}.
\bibitem[{Pham et~al.(2015)Pham, Kheddar, Qammaz and Argyros}]{pham2015towards}
\bibinfo{author}{Pham, T.H.}, \bibinfo{author}{Kheddar, A.},
  \bibinfo{author}{Qammaz, A.}, \bibinfo{author}{Argyros, A.A.},
  \bibinfo{year}{2015}.
\newblock \bibinfo{title}{Towards force sensing from vision: Observing
  hand-object interactions to infer manipulation forces}, in:
  \bibinfo{booktitle}{Proceedings of the IEEE conference on computer vision and
  pattern recognition}, pp. \bibinfo{pages}{2810--2819}.
\bibitem[{Poh et~al.(2010)Poh, McDuff and Picard}]{poh2010advancements}
\bibinfo{author}{Poh, M.Z.}, \bibinfo{author}{McDuff, D.J.},
  \bibinfo{author}{Picard, R.W.}, \bibinfo{year}{2010}.
\newblock \bibinfo{title}{Advancements in noncontact, multiparameter
  physiological measurements using a webcam}.
\newblock \bibinfo{journal}{IEEE transactions on biomedical engineering}
  \bibinfo{volume}{58}, \bibinfo{pages}{7--11}.
\bibitem[{Rao et~al.(2006)Rao, Krishnan and Niculescu}]{rao2006data}
\bibinfo{author}{Rao, R.B.}, \bibinfo{author}{Krishnan, S.},
  \bibinfo{author}{Niculescu, R.S.}, \bibinfo{year}{2006}.
\newblock \bibinfo{title}{Data mining for improved cardiac care}.
\newblock \bibinfo{journal}{ACM SIGKDD Explorations Newsletter}
  \bibinfo{volume}{8}, \bibinfo{pages}{3--10}.
\bibitem[{Salimans et~al.(2016)Salimans, Goodfellow, Zaremba, Cheung, Radford
  and Chen}]{salimans2016improved}
\bibinfo{author}{Salimans, T.}, \bibinfo{author}{Goodfellow, I.},
  \bibinfo{author}{Zaremba, W.}, \bibinfo{author}{Cheung, V.},
  \bibinfo{author}{Radford, A.}, \bibinfo{author}{Chen, X.},
  \bibinfo{year}{2016}.
\newblock \bibinfo{title}{Improved techniques for training gans}, in:
  \bibinfo{booktitle}{Advances in neural information processing systems}, pp.
  \bibinfo{pages}{2234--2242}.
\bibitem[{Samy et~al.(2013)Samy, Huang, Liu, Xu and
  Sarrafzadeh}]{samy2013unobtrusive}
\bibinfo{author}{Samy, L.}, \bibinfo{author}{Huang, M.C.},
  \bibinfo{author}{Liu, J.J.}, \bibinfo{author}{Xu, W.},
  \bibinfo{author}{Sarrafzadeh, M.}, \bibinfo{year}{2013}.
\newblock \bibinfo{title}{Unobtrusive sleep stage identification using a
  pressure-sensitive bed sheet}.
\newblock \bibinfo{journal}{IEEE Sensors Journal} \bibinfo{volume}{14},
  \bibinfo{pages}{2092--2101}.
\bibitem[{Sun et~al.(2019)Sun, Xiao, Liu and Wang}]{sun2019deep}
\bibinfo{author}{Sun, K.}, \bibinfo{author}{Xiao, B.}, \bibinfo{author}{Liu,
  D.}, \bibinfo{author}{Wang, J.}, \bibinfo{year}{2019}.
\newblock \bibinfo{title}{Deep high-resolution representation learning for
  human pose estimation}, in: \bibinfo{booktitle}{CVPR}.
\bibitem[{Tai et~al.(2017)Tai, Yang, Liu and Xu}]{tai2017memnet}
\bibinfo{author}{Tai, Y.}, \bibinfo{author}{Yang, J.}, \bibinfo{author}{Liu,
  X.}, \bibinfo{author}{Xu, C.}, \bibinfo{year}{2017}.
\newblock \bibinfo{title}{Memnet: A persistent memory network for image
  restoration}, in: \bibinfo{booktitle}{Proceedings of the IEEE international
  conference on computer vision}, pp. \bibinfo{pages}{4539--4547}.
\bibitem[{Velardo and Dugelay(2010)}]{velardo2010weight}
\bibinfo{author}{Velardo, C.}, \bibinfo{author}{Dugelay, J.L.},
  \bibinfo{year}{2010}.
\newblock \bibinfo{title}{Weight estimation from visual body appearance}, in:
  \bibinfo{booktitle}{2010 Fourth IEEE International Conference on Biometrics:
  Theory, Applications and Systems (BTAS)}, \bibinfo{organization}{IEEE}. pp.
  \bibinfo{pages}{1--6}.
\bibitem[{Wang et~al.(2004)Wang, Bovik, Sheikh, Simoncelli
  et~al.}]{wang2004image}
\bibinfo{author}{Wang, Z.}, \bibinfo{author}{Bovik, A.C.},
  \bibinfo{author}{Sheikh, H.R.}, \bibinfo{author}{Simoncelli, E.P.}, et~al.,
  \bibinfo{year}{2004}.
\newblock \bibinfo{title}{Image quality assessment: from error visibility to
  structural similarity}.
\newblock \bibinfo{journal}{IEEE transactions on image processing}
  \bibinfo{volume}{13}, \bibinfo{pages}{600--612}.
\bibitem[{Wei et~al.(2016)Wei, Ramakrishna, Kanade and
  Sheikh}]{wei2016convolutional}
\bibinfo{author}{Wei, S.E.}, \bibinfo{author}{Ramakrishna, V.},
  \bibinfo{author}{Kanade, T.}, \bibinfo{author}{Sheikh, Y.},
  \bibinfo{year}{2016}.
\newblock \bibinfo{title}{Convolutional pose machines}.
\newblock \bibinfo{journal}{Proceedings of the IEEE Conference on Computer
  Vision and Pattern Recognition} , \bibinfo{pages}{4724--4732}.
\bibitem[{Wei et~al.(2013)Wei, Li, Cao, Ou and Chen}]{wei2013effective}
\bibinfo{author}{Wei, W.}, \bibinfo{author}{Li, J.}, \bibinfo{author}{Cao, L.},
  \bibinfo{author}{Ou, Y.}, \bibinfo{author}{Chen, J.}, \bibinfo{year}{2013}.
\newblock \bibinfo{title}{Effective detection of sophisticated online banking
  fraud on extremely imbalanced data}.
\newblock \bibinfo{journal}{World Wide Web} \bibinfo{volume}{16},
  \bibinfo{pages}{449--475}.
\bibitem[{Yin and Shi(2018)}]{yin2018geonet}
\bibinfo{author}{Yin, Z.}, \bibinfo{author}{Shi, J.}, \bibinfo{year}{2018}.
\newblock \bibinfo{title}{Geonet: Unsupervised learning of dense depth, optical
  flow and camera pose}, in: \bibinfo{booktitle}{Proceedings of the IEEE
  Conference on Computer Vision and Pattern Recognition}, pp.
  \bibinfo{pages}{1983--1992}.
\bibitem[{Zhang et~al.(2018)Zhang, Tian, Kong, Zhong and
  Fu}]{zhang2018residual}
\bibinfo{author}{Zhang, Y.}, \bibinfo{author}{Tian, Y.}, \bibinfo{author}{Kong,
  Y.}, \bibinfo{author}{Zhong, B.}, \bibinfo{author}{Fu, Y.},
  \bibinfo{year}{2018}.
\newblock \bibinfo{title}{Residual dense network for image super-resolution},
  in: \bibinfo{booktitle}{Proceedings of the IEEE Conference on Computer Vision
  and Pattern Recognition}, pp. \bibinfo{pages}{2472--2481}.
\bibitem[{Zhu et~al.(2017)Zhu, Park, Isola and Efros}]{zhu2017unpaired}
\bibinfo{author}{Zhu, J.Y.}, \bibinfo{author}{Park, T.},
  \bibinfo{author}{Isola, P.}, \bibinfo{author}{Efros, A.A.},
  \bibinfo{year}{2017}.
\newblock \bibinfo{title}{Unpaired image-to-image translation using
  cycle-consistent adversarial networks}, in: \bibinfo{booktitle}{Proceedings
  of the IEEE international conference on computer vision}, pp.
  \bibinfo{pages}{2223--2232}.

\end{thebibliography}
